\documentclass[10pt,journal,onecolumn,compsoc]{IEEEtran}

\ifCLASSOPTIONcompsoc
  \usepackage[nocompress]{cite}
\else
  \usepackage{cite}
\fi
\ifCLASSINFOpdf
  \usepackage[pdftex]{graphicx}
  \graphicspath{{./}{./figures/}{./figures/pinwheel_ablation/}{./figures/mnist_static_components/}{./figures/omniglot_components/}{./figures/caltech_components/}}
\else
\fi
\usepackage{amsmath}
\allowdisplaybreaks

\usepackage{amsmath,amsfonts,bm}

\def\eqref#1{equation~\ref{#1}}

\def\1{\bm{1}}

\DeclareMathAlphabet{\mathsfit}{\encodingdefault}{\sfdefault}{m}{sl}
\SetMathAlphabet{\mathsfit}{bold}{\encodingdefault}{\sfdefault}{bx}{n}

\newcommand{\E}{\mathbb{E}}

\usepackage{amssymb}
\usepackage{color} 
\usepackage{colortbl}
\usepackage{hyperref}
\usepackage{multirow}
\usepackage{caption}
\usepackage{subcaption}
\usepackage{tikz}
\usepackage{comment}
\usepackage{url}
\usepackage{booktabs}

\newcommand{\mb}[1]{\mathbf{#1}}
\newcommand{\rulesep}{\unskip\ \vrule\ }
\newcommand{\pms}[1]{\tiny{$\pm$ #1}}
\newcommand*\circled[1]{\tikz[baseline=(char.base)]{
                \node[shape=circle,draw,inner sep=2pt] (char) {#1};}}
\newcommand{\fid}[1]{Fr\'{e}chet inception distance{#1}}
\newcommand{\norm}[1]{\left\lVert#1\right\rVert}

\renewcommand{\eqref}[1]{(\ref{#1})}

\definecolor{Gray}{gray}{0.9}

\begin{document}
%
\title{Cauchy-Schwarz Regularized Autoencoder}
%
%
%
%

\author{Linh~Tran,~\IEEEmembership{Member,~IEEE,}
        Maja~Pantic,~\IEEEmembership{Fellow,~IEEE,}
        and~Marc~Peter~Deisenroth
\IEEEcompsocitemizethanks{\IEEEcompsocthanksitem L. Tran is with the Department
of Computing, Imperial College London, United Kingdom.\protect\\
E-mail: linh.tran@imperial.ac.uk
\IEEEcompsocthanksitem M. Pantic is with the Department of Computing, Imperial College London, United Kingdom and Facebook AI Applied Research, United Kingdom.
\IEEEcompsocthanksitem M. P. Deisenroth is with the Department of Computer Science, University College London, United Kingdom.}
}

\IEEEtitleabstractindextext{%
\begin{abstract}
    Recent work in unsupervised learning has focused on efficient inference and learning in latent variables models.
    Training these models by maximizing the evidence (marginal likelihood) is typically intractable.
    Thus, a common approximation is to maximize the Evidence Lower BOund (ELBO) instead.
    Variational autoencoders (VAE) are a powerful and widely-used class of generative models that optimize the ELBO efficiently for large datasets.
    However, the VAE's default Gaussian choice for the prior
    imposes a strong constraint on its ability to represent the true posterior, thereby degrading overall performance.
    A Gaussian mixture model (GMM) would be a richer prior, but  cannot be handled efficiently within the VAE framework because of the intractibility of the Kullback-Leibler divergence for GMMs.
    We deviate from the common VAE framework
    in favor of one with an analytical solution for Gaussian mixture prior.
    To perform efficient inference for GMM priors,
    we introduce a new constrained objective based on the Cauchy-Schwarz divergence, which can be computed analytically for GMMs.
    This new objective allows us to incorporate richer, multi-modal priors
    into the autoencoding framework.
    We provide empirical studies on a range of datasets and show that our objective improves upon variational auto-encoding models in density estimation, unsupervised clustering, semi-supervised learning, and face analysis.
\end{abstract}

}

\maketitle

\IEEEdisplaynontitleabstractindextext

%
\IEEEpeerreviewmaketitle

\ifCLASSOPTIONcompsoc
\IEEEraisesectionheading{\section{Introduction}\label{sec:introduction}}
\else
\section{Introduction}
\label{sec:introduction}
\fi
In recent years, generative models have made 
have made remarkable progress in learning complex, high-dimensional distributions. 
Particularly deep generative models can model highly complex datasets, including natural images and speech
\cite{kingma2013auto,rezende2014stochastic,goodfellow2014generative,radford2015unsupervised,oord2016pixel,arjovsky2017wasserstein}.
Variational autoencoders (VAEs), a class of deep generative models, are likelihood-based models which allow to model high-dimensional distributions in a probabilistic fashion.
VAEs approximate the true data distribution by maximizing a tractable lower bound on the marginal likelihood (evidence), also known as the evidence lower bound (ELBO).
This is equivalent to minimizing the expected negative log-likelihood and Kullback-Leibler (KL) divergence between approximate posterior and prior.
With the introduction of VAEs~\cite{kingma2013auto,rezende2014stochastic}, it has become a popular choice of framework for generative modeling. It is based on well-established theory, its training is simpler and more stable than Generative Adversarial Networks (GANs)~\cite{goodfellow2014generative} and its sampling is more efficient than autoregressive models.

Although there has been a surge of work focused to applying VAEs to image generation tasks and improving its encoder–decoder architectures, the VAE framework is still far from learning generative mechanism for real-world scenarios.
One of the main challenges is that VAE samples tend to be blurry.
Blurry samples can be attributed to the overly simplistic distribution for the prior~\cite{chen2016variational,nalisnick2016approximate,nalisnick2017stick} or posterior~\cite{rezende2015variational} and the overregularization through the KL term of VAE objective~\cite{higgins2016beta}.
As a result several mechanisms focusing on increasing the expressiveness of the variational posterior density
\cite{rezende2015variational,salimans2015markov,tran2015variational,nalisnick2016approximate,
gregor2016towards,kingma2016improved,tomczak2016improving,van2018sylvester} have been proposed.
Recent work~\cite{johnson2016composing,hoffman2016elbo} have shown that the prior plays an essential part balancing the performance of the probabilistic encoder and decoder.
In the common case of VAE, the choice of a simple prior, e.g., a Gaussian prior with zero mean and unit diagonal covariance, leads to poor generalization due to overregularization of the approximate posterior.
In particular, several approaches use \textit{Gaussian mixture models} (GMMs) as priors for VAEs
\cite{dilokthanakul2016deep,zheng2016variational,tomczak2016improving} to increase model capability.
However, these approaches do not allow for closed-form optimization. Estimates with low number of Monte Carlo samples can lead to high variance and thus unstable training. Estimates with higher number of Monte Carlo samples are expensive to compute.

In our work, we addresses the aforementioned challenges by changing the VAE framework into a generative model which is better suited for GMMs.
We propose an approach which focuses on the loss-function formulation of the VAE: 
We substitute the VAE objective function with an explicit regularization scheme based on the Cauchy-Schwarz (CS) divergence.
This loss-formulation is no longer a lower bound.
Nevertheless, this new approach allows us to use GMMs as effective priors since the Cauchy-Schwarz divergence between GMMs can be computed analytically.

Our main contribution is the Cauchy-Schwarz regularized autoencoder (CSRAE) framework for generative modeling and the introduction of a new objective that provides a closed-form solution on the divergence between GMM prior and variational posterior. Compared to existing variational models, we improve on a range of computer vision based tasks: 1) density estimation, 2) unsupervised clustering, 3) semi-supervised learning and 4) face analysis.
\section{Background}
\label{sec:background}
In a typical unsupervised learning setting, we are often interested in learning compact representations from unlabelled data. For that, we assume to have a dataset $D$ and define $p_D(x)$ be the empirical data distribution defined over i.i.d. samples of $D$.
Latent variable models defines a joint (parameterized) probability distribution $p_{\theta}(x, z)$ with $x$ being the observed variables and $z$ being (a set of) latent variables.
In the Bayesian setting, the joint probability distribution  $p(x, z)$ is assumed to have a prior $p(z)$ and a likelihood $p_{\theta}(x|z)$ and the target distribution is the posterior $p(z|x)$.
For simplicity, the prior $p(z)$ is usually a Gaussian or uniform prior which enables efficient optimization.
%
\subsection{Variational Autoencoders}
\label{subsec:vae}
In the typical VAE setting~\cite{kingma2013auto}, the posterior distribution $p(z|x)$ is intractable because the log-marginal likelihood $p(x) = \int p_{\theta}(x,z)dz$ cannot be computed analytically. Therefore, it is common to introduce a parametric approximation $q_{\phi}(z|x)$ to the intractable posterior and minimize the KL divergence between $q$ and the true posterior $p(z|x)$. This is equivalent to maximizing a lower bound $\mathcal L_{\textrm{ELBO}}$ on the log-marginal likelihood, which is given by
\begin{align}
    \begin{split}
    \log p(x) \ge & \; \E_{q_{\phi}}[\log p_{\theta}(x|z)]  - \textrm{D}_{\textrm{KL}}(q_{\phi}(z|x) \parallel p(z)) \\
    =: & \; \mathcal{L}_{\textrm{ELBO}}(x;\theta,\phi),\label{eq:elbo}
    \end{split}
\end{align}
where $\theta$ are the model parameters of $p$ and $ \phi$ are the variational parameters of $q$. 

There are various ways to optimize the lower bound~\eqref{eq:elbo}.
For continuous latent variables $z$ it can be done efficiently through a reparameterization of the approximate posterior
$q_{\phi}(z|x)$~\cite{kingma2013auto,rezende2014stochastic}.
If both the prior $p(z)$ and the variational approximation $q_{\phi}(z|x)$  are Gaussian, the KL term in~\eqref{eq:elbo} can be computed in closed form~\cite{kingma2013auto}. In the common case $p(z)$ is set to be a mean-field approximation (factorizing Gaussian). 
The mean-field approximation has the advantage of being efficient because it assumes all latent variables to be independent and thus, it simplifies the derivations
However, this simplified assumption also leads over-regularization.

Although VAEs are widely used for large-scale approximations, it has been observed that VAEs lack in sample quality because the optimized model simplifies the posterior distribution. 
Maximizing~\eqref{eq:elbo} with respect to the variational parameters $\phi$ amounts to
minimizing the KL divergence $\textrm{D}_{\textrm{KL}}(q_{\phi}(z|x) \parallel p(z|x))$ between the approximate posterior and the true posterior. 
The mean-field variational family is problematic, as the log-marginal likelihood $\log p_D(x)$ can only be optimized to the extent we are able to approximate the true posterior using this restricted variational family.
As a result, considering  richer families of approximate posteriors\cite{rezende2015variational,kingma2016improved,kucukelbir2015automatic} and richer families of priors \cite{tomczak2016improving,kuznetsov2019prior,chen2016variational,nalisnick2016approximate,ghosh2019variational} have been proposed to improve VAE-based models.

From a maximum likelihood perspective, VAE can be seen as an approach to maximize the likelihood of the model. This equates to the first term of~\eqref{eq:elbo}. The maximization is regularized by a KL term between approximate posterior and prior (second term of~\eqref{eq:elbo}). Motivated by this loss-centric view to minimize the divergence between approximate posterior and prior we diverge from the variational inference principle and propose a constrained optimization objective with the Cauchy-Schwarz divergence to compute the divergence between GMMs analytically.
\subsection{Cauchy-Schwarz divergence}
\label{subsec:csdiv}
Based on the Cauchy-Schwarz inequality
$$ \left\lVert x \right\rVert^2 \left\lVert y \right\rVert^2 \ge (x^T y)^2,$$
the Cauchy-Schwarz divergence~\cite{principe2010information} is defined as 
\begin{align}\label{eq:dcs}
\textrm{D}_{\textrm{CS}} (q(x) \parallel p(x)) = & \; - \log
\frac{\smallint q(x) p(x) dx}{\sqrt{\smallint p(x)^2 dx \smallint q(x)^2 dx}}\\
= &\; - \log
\smallint q(x) p(x) dx 
 + 0.5 \log \smallint p(x)^2 dx \nonumber \\
 & \;  + 0.5 \log \smallint q(x)^2 dx.
\label{eq:dcs_general}
\end{align}
This is a symmetric metric for any two probability density functions $p$ and $q$, such that
$0 \le \textrm{D}_{\textrm{CS}} < \infty$, where the minimum is obtained if
and only if $q(z) = p(z)$.
Principe et al.~\cite{principe2010information} also show empirical results that indicate that the CS divergence can be considered an approximation to the Kullback-Leibler divergence.

\textbf{Analytical solution for mixture of Gaussians} The Kullback-Leibler divergence can only be computed in closed form for Gaussians, but not for the more versatile class of Gaussian mixtures.
However, the Cauchy-Schwarz divergence can be computed in closed form for Gaussian mixtures~\cite{kampa2011closed}, a property we will exploit in this paper.
For example, let
$q(x) = \sum_{n=1}^N w_{n} \mathcal{N}(x | \mu_{n}, \sigma_{n}^2)$ and $p(x) = \sum_{m=1}^M v_{m} \mathcal{N}(x | \nu_{m}, \tau_{m}^2)$ be two mixture-of-Gaussian distributions with different parameters and different numbers of mixture components.
Applying that to the three $\log$ terms of~\eqref{eq:dcs_general} separately,
the closed-form expression for the Cauchy-Schwarz divergence between $q$ and $p$ translates into
\begin{align}
\begin{split}
\mathrm{D}_{\mathrm{CS}} =
& \; - \;\log \Big(\sum_{n=1}^N \sum_{m=1}^M w_{n} v_{m} z_{n,m} \Big) \\
& \; + 0.5 \log \Big( \sum_{n, n'}^N  w_{n} w_{n'} z_{n,n'} \Big) \\
& \;+ 0.5 \log \Big( \sum_{m, m'}^M  v_{m} v_{m'} z_{m,m'} \Big),
\end{split}
\end{align}
where we define $z_{n,m}$, $z_{m,m'}$ and $z_{n,n'}$ as 
\begin{align}
z_{n,m} = & \; \mathcal{N}(\mu_{n} | \nu_{m}, \sigma^2_{n} + \tau^2_{m})\\
z_{m,m'} = & \; \mathcal{N}(\nu_{m} | \nu_{m'}, 2 \tau^2_{m'})\\
z_{n,n'} = & \; \mathcal{N}(\mu_{ n} | \mu_{n'}, 2\sigma^2_{n'}).
\end{align}
A detailed derivation of the analytical form is given in the Appendix (1.2).
\section{Cauchy-Schwarz Regularized Autoencoder}\label{sec:CSRAE}
We consider the objective to maximize the log-marginal likelihood of the model, i.e.,
\begin{align}
\max_{\theta} \; \E[\log p_{\theta}(x)]= \max_{\theta} \; \E_{p_D(x)} [\log \E_{p(z)}[p_{\theta}(x|z)]],
\end{align}
where the expectation w.r.t. $p_D(x)$ is approximated using a sample average over the training data $D$. By using the Jensen’s inequality we can obtain a lower bound to the log-marginal likelihood as
\begin{align}
\log p_\theta(x) = \log \E_{p(z)}[p_{\theta}(x|z)] \ge \E_{p(z)}[\log p_{\theta}(x|z)].
\end{align}
This objective does not regularize the encoding distribution. Sampling from the model once training is completed is difficult and thus, maximizing this objective alone can lead to poor generalization. 
Similarly as in the VAE framework, we define a mapping $q_{\phi}(z|x)$ which transforms the input $x$ to (probabilistic) features $z$.
However, we do not treat $q(z|x)$ as an approximate posterior to the true posterior $p(z|x)$.
Rather, we match the approximate posterior to the prior to enforce a way of sampling from the generative model $p_{\theta}(x|z)p(z)$. By adding a regularization $R$ we penalize any deviation between  $q_{\phi}(z|x)$ from $p(z)$. 
Ideally, this regularization is a metric function for which $R > 0$ when $q \neq p$ and $R=0$ if and only if $q=p$.
%
\begin{align}
\begin{split}
\max_{\theta, \phi} \E_{p_D(x)} \E_{q_{\phi}(z|x)}[\log p_{\theta}(x|z)]\\
\textrm{ subject to } 0 \le R(q_{\phi}) < \epsilon.
\end{split}\label{eq:Rconstraint}
\end{align}
In this formulation $\epsilon$ specifies the magnitude of the applied constraint. If $R$ is defined as KL divergence, we have the original ELBO formulation~\eqref{eq:elbo}. We divert from this principle and use the Cauchy-Schwarz divergence for regularization to match an approximate posterior to a prior. The advantage is that we can use GMMs, a powerful family of distributions, and calculate the divergence analytically. This enables both prior and approximate posterior to be more flexible.
\begin{align}
\begin{split}
 \underset{\theta, \phi}{\max} \; \E_{p_{D}(x)}\Big[ \E_{q_{\phi}(z|x)}[\log p_{\theta}(x|z)]\Big]\\
\text{ subject to } \mathrm{D}_{\mathrm{CS}}(q_{\phi}(z|x) \parallel p(z)) < \epsilon.
\label{eq:CS:constrainedopt}
\end{split}
\end{align}
Re-writing~\eqref{eq:CS:constrainedopt} as a Lagrangian under the KKT conditions 
\cite{karush1939minima,kuhn1951nonlinear}, we obtain
\begin{align}
\begin{split}
\mathcal{F}(x;\theta, \phi, \lambda) =& \; \E_{q_{\phi}(z|x)}[\log p_{\theta}(x|z)] \\
& \; - \lambda (\mathrm{D}_{\mathrm{CS}}(q_{\phi}(z|x) \parallel p(z)) - \epsilon),
\end{split}\label{eq:CS:lagrangian}
\end{align}
where the KKT multiplier $\lambda$ is the regularization coefficient that ensures that the posterior distribution is close to the prior $p(z)$. According to the complementary slackness
of the KKT condition and since $\alpha, \epsilon \ge 0$,~\eqref{eq:CS:lagrangian} can be re-written as
\begin{align}
\mathcal{F}(x;\theta, \phi, \lambda) \ge& \; \E_{q_{\phi}(z|x)}[\log p_{\theta}(x|z)]  - \lambda \mathrm{D}_{\mathrm{CS}}(q_{\phi}(z|x) \parallel p(z))\nonumber \\
=: & \; \mathcal{L}_{\textrm{CSRAE}}(x; \theta, \phi, \lambda).
\label{eq:csrae}
\end{align}
Decomposing the proposed objective $\mathcal{L_{\textrm{CSRAE}}}(x;\theta, \phi, \lambda)$ yields
\begin{align}
    \begin{split}
         \mathcal{L}_{\textrm{CSRAE}} = & \; \underbrace{\log p(x)}_{\substack{\text{log-marginal}\\ \text{likelihood}}} - \underbrace{\textrm{D}_{\textrm{KL}}(q(z|x) \parallel p(z | x))}_{\substack{\text{KL divergence between}
         \\
         \text{posterior and approx. posterior}}} \\
         & + \; \underbrace{\textrm{D}_{\textrm{KL}}(q(z|x) \parallel p(z))}_{\substack{\text{KL divergence between}
         \\
         \text{approx. posterior and prior}}}
         -\lambda \cdot \underbrace{\textrm{D}_{\textrm{CS}}(q(z|x) \parallel p(z))}_{\substack{\text{CS divergence between}
         \\
         \text{approx. posterior and prior}}}. 
         \end{split}\label{eq:decompose-csrae}
\end{align}
Maximizing $\mathcal{L}_{\textrm{CSRAE}}$ is equivalent to maximizing the log-marginal likelihood, mimimizing the KL divergence between approximate posterior and true posterior and minimizing/maximizing the divergence between approximate posterior and prior.

\subsection{Relation to \texorpdfstring{$\beta$}--VAE and rate-distortion theory}

 $\mathcal{L}_{\textrm{CSRAE}}$ is exactly the log-marginal likelihood if $q(z|x) = p(z | x) = p(z)$. 
 The hyperparameter $\lambda$ determines the pressure applied to the regularization during training which encourages different degrees of how well the approximate posterior matches the prior.
 This improves on sampling from the autoencoding model, but may also decrease the expected log likelihood (\textit{reconstruction}) under the model.
 This behaviour is similar to the one of $\beta$-VAE~\cite{higgins2016beta}. $\beta$-VAE~\cite{higgins2016beta} modified the VAE framework~\cite{kingma2013auto,rezende2014stochastic} and introduces  $\beta$ to adjusting the KL term for better controlling 
 the degree of disentanglement.
 It has a similar objective, which we can be similarly decomposed as in~\eqref{eq:decompose-csrae}:
\begin{align}
\begin{split}
\mathcal{L}_{\beta} = & \; \E_{q_{\phi}(z|x)}[\log p_{\theta}(x|z)]  - \beta \mathrm{D}_{\mathrm{KL}}(q_{\phi}(z|x) \parallel p(z))
\end{split}\\
\begin{split}
= & \; \underbrace{\log p(x)}_{\substack{\text{log-marginal}\\ \text{likelihood}}} -  \underbrace{\textrm{D}_{\textrm{KL}}(q(z|x) \parallel p(z | x))}_{\substack{\text{KL divergence between} \\
 \text{posterior and approx. posterior}}} \\
 & + \; (\beta - 1) \cdot \underbrace{\textrm{D}_{\textrm{KL}}(q(z|x) \parallel p(z))}_{\substack{\text{KL divergence between}\\
 \text{approx. posterior and prior}}}.
\end{split}
\end{align}
Similar to our approach, the $\beta$-VAE puts a contraint on the similarity between the approximate posterior and the prior through a regularized KL divergence. Both the proposed CSRAE objective  $\mathcal{L}_{\textrm{CSRAE}}$ and the $\beta$-VAE objective~\cite{higgins2016beta} $\mathcal{L_{\beta}}$ are more general optimization criteria that are not always a lower bound on the log-marginal likelihood. For $\beta$-VAE $\mathcal{L_{\beta}}$ is a lower bound with $\beta \ge 1$. For CSRAE we have a lower bound on the log-marginal likelihood if $\lambda  \textrm{D}_{\textrm{CS}}(q(z|x) \parallel p(z)) \ge \textrm{D}_{\textrm{KL}}(q(z|x) \parallel p(z))$. However, the inequality cannot be solved analytically for $\lambda$. Similar to  $\beta$-VAE~\cite{higgins2016beta}, we believe there is a trade-off between the quality of the reconstruction and mis-match between prior and approximate posterior. The greater $\lambda$ the closer the approximate posterior is to the prior which improves sampling from the prior. However, this can also degrade the reconstruction due to the pressure of $\lambda$. We treat $\lambda$ as a hyperparameter and optimize for a Pareto optimal solution between reconstruction and constraint, i.e., we perform model selection according to the model achieving $\min \Big(\E_{q_{\phi}}[\log p_{\theta}(x|z)] + \textrm{D}_{\textrm{CS}}(q_{\phi}(z|x) \parallel p(z))\Big)$ on the validation set.

We can also take an information theoretic perspective as was done in~\cite{higgins2016beta,burgess2018understanding,alemi2018fixing} for the KL divergence and which also applies to the Cauchy-Schwarz divergence. Any auto-encoding model can be seen as a communication channel trying to transmit data. The approximate posterior $q(z|x)$ can be considered independent noise channels $z_i$ as it has diagonal covariance matrix and thus is factorized. From this perspective, the CS divergence $\textrm{D}_{\textrm{CS}}(q(z|x) \parallel p(z))$ can be seen as the upper bound of number of information required to represent data. If the Cauchy-Schwarz divergence is zero, then each channel $z_i$ has zero capacity and thus cannot transmit any data in the channel.
The only way to increase the capacity of each channel is to vary the posterior means or decrease the posterior variance.
Both ways increase the CS divergence and thus also the capacity.


%
\subsection{Mixture Cauchy-Schwarz regularized autoencoder}
One of our main motivations for the proposed constrained optimization objective is to use the Cauchy-Schwarz divergence for a mixture of Gaussian prior. Similar to the constrained problem defined in~\eqref{eq:csrae} we now define Mixture Cauchy-Schwarz regularized autoencoder (MixtureCSRAE) using mixture of Gaussian prior. We consider the inference model
\begin{align}
q_{\phi}(z|x) =& \; \mathcal{N}(z|\mu_{\phi}(x), \textrm{diag}(\sigma^2_{\phi}(x))) \label{eq:q_zx_mixture}
\end{align}
and the  generative model
\begin{align}
p(z) = & \; \frac{1}{K} \sum_{k=1}^{K} \mathcal{N}(z|\mu_{k}, \textrm{diag}(\sigma^2_{k})),\label{eq:p_z}\\
p_{\theta}(x|z) = & \; \textrm{Bernoulli}(f_{\theta}(z)).
\end{align}
Similar to the VAE framework we assume the approximate posterior is Gaussian distributed as defined in~\eqref{eq:q_zx_mixture}. We parameterize both mixture means $\mu_{\phi}(\cdot)$ and variances $\sigma^2_{\phi}(\cdot)$ through an encoding neural network with $x$ as input. A prior put on the approximate posterior is defined as a $K$-component mixture of Gaussians, c.f.~\eqref{eq:p_z}. 
The decoder is parameterized by a neural network with weights $\theta$
It uses an appropriate likelihood for the respective data used for optimization, e.g., a Bernoulli likelihood for binary data or the Gaussian likelihood for continuous data.
Putting everything together we get the  constrained optimization objective
\begin{align}
    \begin{split}
        \mathcal{L}_{\textrm{MixtureCSRAE}} = & \; \E_{q_{\phi}(z|x)}[\log p_{\theta}(x|z)]\\
        & \; - \lambda \textrm{D}_{\textrm{CS}}(q(z|x)\parallel p(z))
    \end{split}\\
    \begin{split}\label{eq:mixturecsrae}
        = & \; \E_{q_{\phi}(z|x)}[\log p_{\theta}(x|z)] \\
        & \; + \lambda \log \Big(
        \sum_{k}^{K} \mathcal{N}\big(\mu_{\phi} | \mu_{k, \psi}, \textrm{diag}(\sigma^2_{\phi} + \sigma^2_{k, \psi})\big)\Big)\\ 
        & \; - \lambda \log \Big(\sum^K_{k, k'} \mathcal{N}(\mu_{k, \psi}|\mu_{k', \psi}, \textrm{diag}(2\sigma^2_{k',\psi}))\Big)\\
        & \; + \lambda D \log (2\sigma_{\phi} \sqrt{\pi}) - \lambda\log K 
\end{split}
\end{align}

The first term of \eqref{eq:mixturecsrae} represents the reconstruction error. The remaining terms optimize the approximate posterior and the prior with $\lambda$ controlling its degree. The second term is maximized if the approximate posterior mean $\mu_{\phi}$ is close to one of the prior means $\mu_{k,\psi},\; k=1,\ldots, K$. The third term penalizes mean priors $\mu_{k,\psi}$ if they are close together and hence, avoids clusters degrading to one cluster. The second last term penalizes the loss term if the approximate posterior increases. The last term can be considered constant w.r.t. to number of clusters. The full derivation of the objective can be found in the Appendix (1.3).

\textbf{Choice of prior} The mixture means and variances of $p(z)$ are easier to set for lower-dimensional cases. However, setting mixture parameters in high dimensions is non-trivial. Therefore, similar to VampPriorVAE~\cite{tomczak2018vae}, we learn the parameters of $p(z)$ through a neural network. 
There are two kinds of priors we will investigate for our evaluation:
\begin{enumerate}
    \item mixture of Gaussians prior (MoGPrior):
    \begin{align}
        p(z) = \frac{1}{K} \sum_{k=1}^{K} \mathcal{N}(z|\mu_{k}, \textrm{diag}(\sigma^2_{k}))
    \end{align}
    \item variational mixture of posteriors prior (VampPrior) which uses the generative decoder to learn representative pseudo-inputs ($u_{k}$ of the data:
    \begin{align}
        p(z) = \frac{1}{K} \sum_{k=1}^{K} q_{\phi}(z|u_{k}).
    \end{align}
\end{enumerate}
The first prior requires more parameters for tuning as it is defined in a general way and there are multiple ways for implementation. The VampPrior which was also used in  VampPriorVAE~\cite{tomczak2018vae} is simple to apply as it uses the encoder to output mixture means and variances. However, it is also prone to overfitting.
\subsection{Semi-Supervised Learning}\label{sec:ssl}
In the semi-supervised setting we consider a labelled dataset $D = \{(x_1, y_1), \ldots,(x_N , y_N )\}$, composed of observations $x_i$ and corresponding class labels $y_i \in \{1, \ldots, K\}$  where $K$ equals to the total number of classes. 
In this setting, only a small part of the dataset is labelled. Therefore, we split the dataset $D$ into two subsets: $D_{l}$ for labelled training and $D_{ul}$ for unlablled training with $D_{l} \cup D_{ul} = D$ and $D_{l} \cap D_{ul} = \emptyset$.
The labels are used to condition the probabilistic model. In the unsupervised subsetting, we treat the label as a second latent variable. We refer to Kingma et al.~\cite{KingmaMRW14} for the generative and inference model.
%

%
%
For this semi-supervised model we have to consider two cases, the supervised and unsupervised objective. In the supervised case, the labels are observed and we can perform inference on $z \sim q_{\phi}(z|x)$ only. Thus, we have the following constrained optimization objective:
\begin{align}
\begin{split}
    \max_{\theta,\phi} \mathbb{E}_{p_{D_l}}\Big[\mathbb{E}_{q_{\phi}(z|x)}(\log p_{\theta}(x|z,y)\Big] \\
    \textrm{subject to }0 <  \mathrm{R}(q_{\phi}(z|x) \parallel p(z)) < \epsilon_1\\
    \textrm{subject to }0 < \mathrm{R}(q_{\phi}(y|x) \parallel p(y)) < \epsilon_2.
\end{split}
\end{align}
As defined before in~\eqref{eq:Rconstraint}, $R$ penalizes deviation from approximate posterior and prior. For the latent variable $z$ we chose $R$ to be the Cauchy-Schwarz divergence. This allows for an analytical solution of mixture of Gaussians. For the (latent) variable $y$ we chose $R$ to be a KL divergence. This choice allows for an analytical solution of Categorical variables. Further, it is similar to the choice made by Kingma et al.~\cite{KingmaMRW14} and thus, allows for a fair comparison in the evaluation as the objective only differs in the choice of divergence and prior for $z$. As a result, we have the following objective for labelled observations:
\begin{align}
    \begin{split}
        \mathcal{L}_{\textrm{ssl}}(x, y) = & \;\mathbb{E}_{q_{\theta}(z|x)}[\log p_{\theta}(x|z, y)]\\
        & \; - \lambda \textrm{D}_{\textrm{CS}}(q(z|x)|p(x))\\
        & \; - \beta \textrm{D}_{\textrm{KL}}(q(y|x)|p(y)).
    \end{split}
\end{align}
In the unsupervised case, the label is treated as a latent variable over which we marginalize for inference.The resulting objective for handling data points with an unobserved label $y$ is
\begin{align}
    \mathcal{U}_{\textrm{ssl}}(x) = \sum_y \mathcal{L}_{\textrm{ssl}}(x, y).
\end{align}
The final objective function is
\begin{align}
    \begin{split}
        \mathcal{J} = &\; \mathbb{E}_{x,y \sim p_{D_l}}[\mathcal{L}_{\textrm{ssl}}(x, y)] + \mathbb{E}_{x \sim p_{D_u}}[\mathcal{U}_{\textrm{ssl}}(x)]\\ &\; + \alpha \mathbb{E}_{x,y \sim p_{D_l}}[q_{\phi}(y|x)].
    \end{split}
\end{align}
This objective combines both supervised and unsupervised objectives as defined before and adds a classification as done in~\cite{KingmaMRW14}. This ensures that the overall objective and in particular the distribution $q_{\phi}(y|x)$  also learn from from the labelled data.
%
%

%
%
\textbf{Multi-output labels} When learning multi-output labels, we have labels $y \in \{0, 1\}^L$ where $L$ is the number of ouputs. We assume a factorized prior for $y$ given by $p(y)=\prod_{i=1}^L p(y^{i}), \; p(y^i) = \textrm{Cat}(y|\pi)$. If using the semi-supervised model described before, the unlabelled objective requires marginalizing over all possible label classes. In the setting of a label with $L$ outputs $y \in \{0, 1\}^L$ marginalizing over all possible label combinations equals to $2^{L}$ possible combinations. Instead we approximate $y$ by sampling from the Gumbel-Softmax distribution~\cite{JangGP17}. 
%
\section{Evaluation}\label{sec:eval}
Our primary goal is to quantitatively and qualitatively assess properties of the newly proposed CSRAE and Mixture CSRAE. Further, we would like to answer the following questions for the evaluation:
\begin{enumerate}
    \item Do CSRAE and MixtureCSRAE improve on sample quality compared to VAE and its variants?
    \item Can the learned latent embedding used for clustering related tasks?
    \item Can CSRAE and MixtureCSRAE be applied for semi-supervised learning and face-related tasks?
\end{enumerate}

%
\subsection{Experimental setups}\label{sec:setup}
\begin{figure*}[t!]
\centering
\subcaptionbox{KL divergence with univariate Gaussian}{\includegraphics[width=0.35\linewidth]{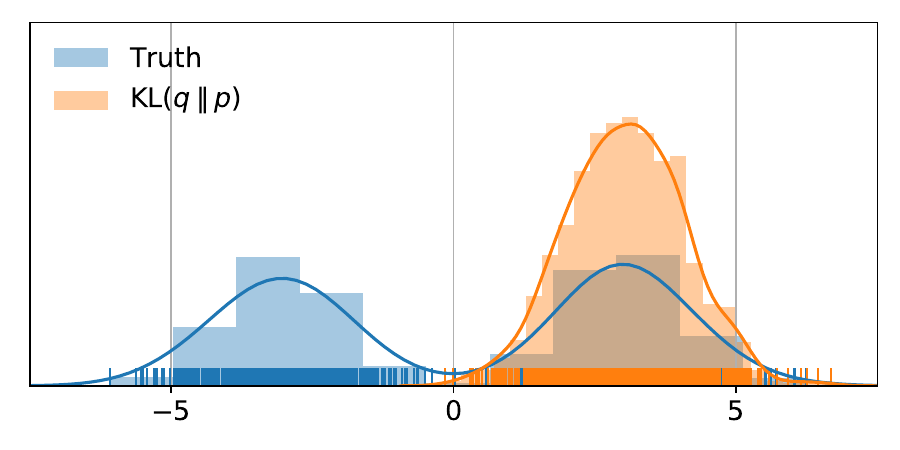}\label{fig:kl_vs_cs-kl}}%
\subcaptionbox{CS divergence with mixture of two Gaussians}{\includegraphics[width=0.35\linewidth]{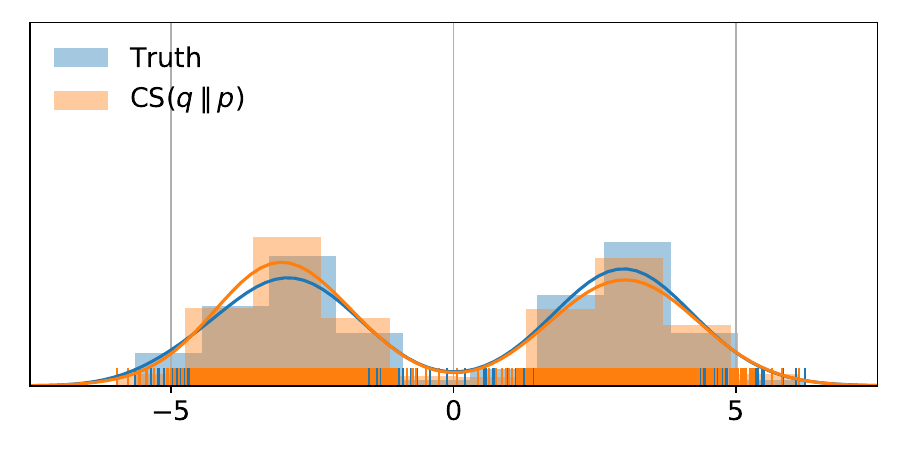}\label{fig:kl_vs_cs-cs}}
\caption{The true posterior is a mixture of two univariate Gaussians (blue). The approximate posterior is a univariate Gaussian for KL and a mixture of univariate Gaussians on the right (orange). The KL approximation is on the left (a), the CS approximation on the right (b).}
\label{fig:toy_mixture}
\end{figure*}
\begin{figure*}[ht]
    \centering
    \subcaptionbox{Prior}{\includegraphics[width=0.136\textwidth]{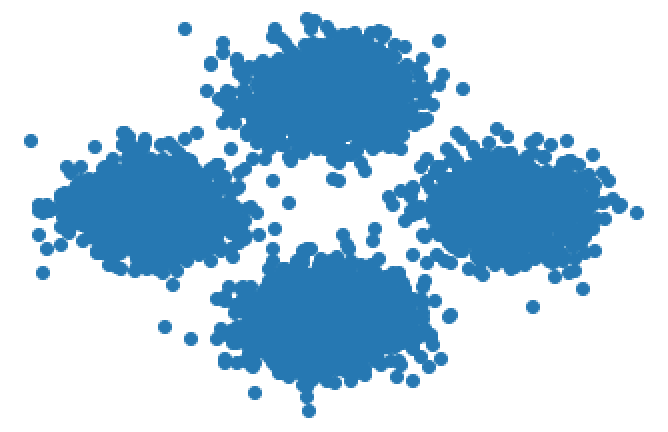}}
    \rulesep
    \hfill
    \subcaptionbox{$\lambda=0.5$}{\includegraphics[width=0.136\textwidth]{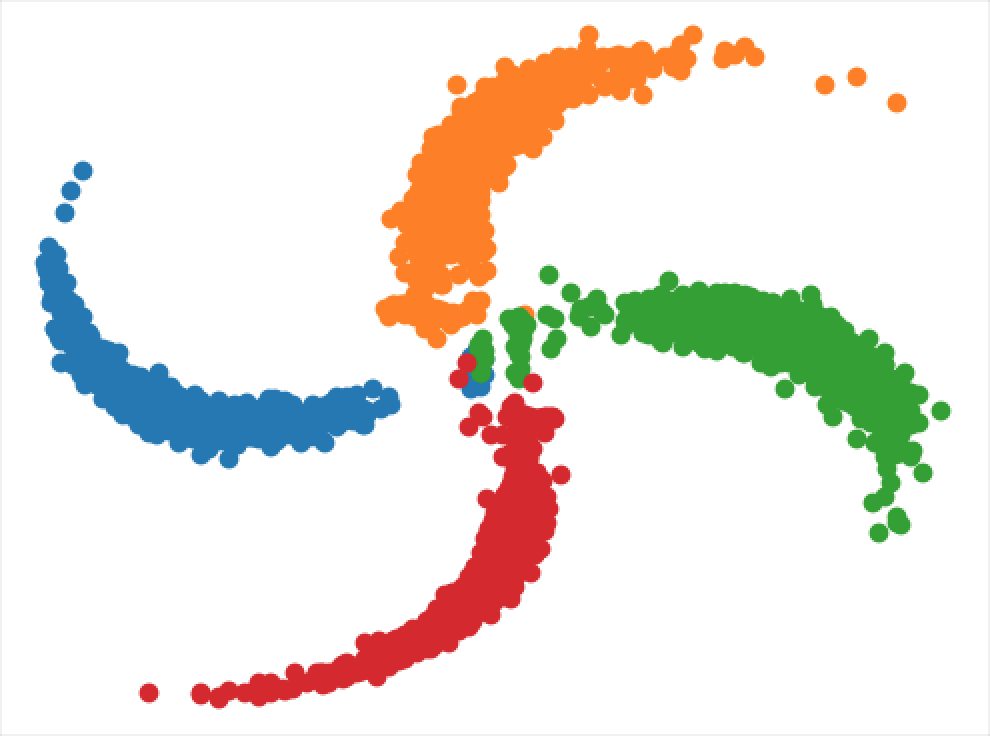}}
    \hfill
    \subcaptionbox{$\lambda=1.0$}{\includegraphics[width=0.136\textwidth]{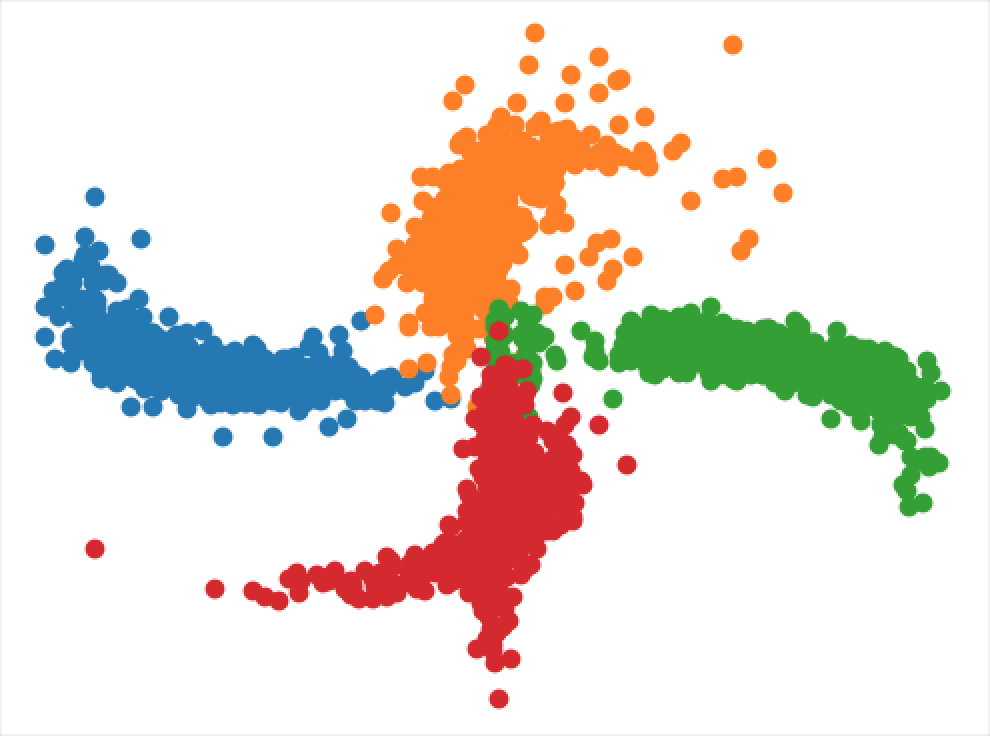}}
    \hfill
    \subcaptionbox{$\lambda=2.5$}{\includegraphics[width=0.136\textwidth]{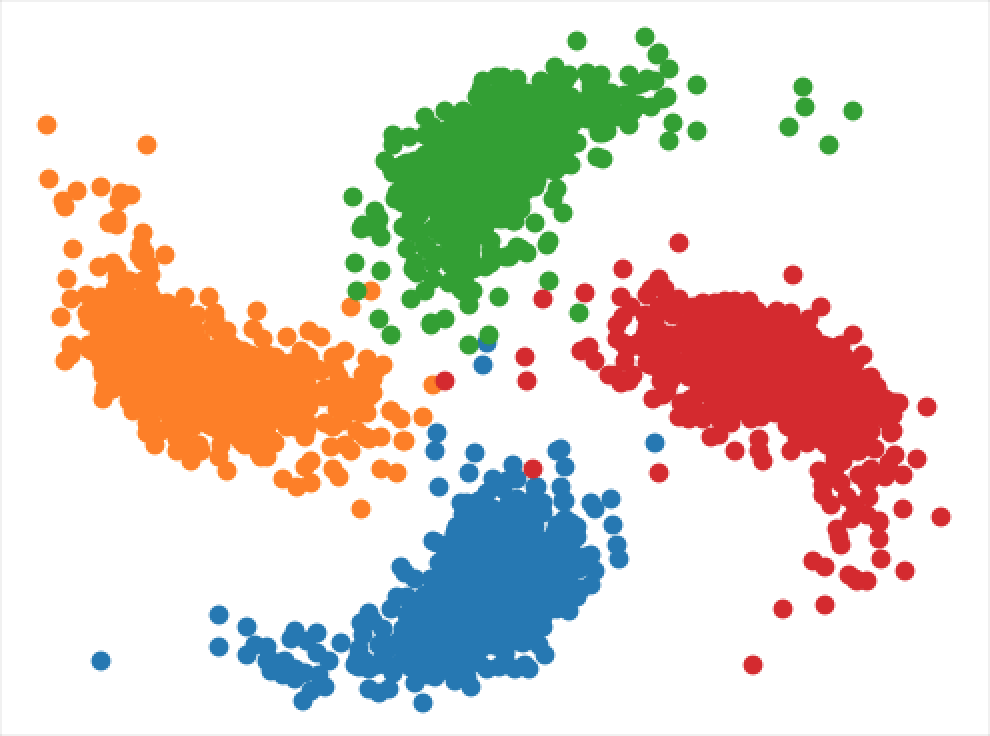}}
    \hfill
    \subcaptionbox{$\lambda=5.$}{\includegraphics[width=0.136\textwidth]{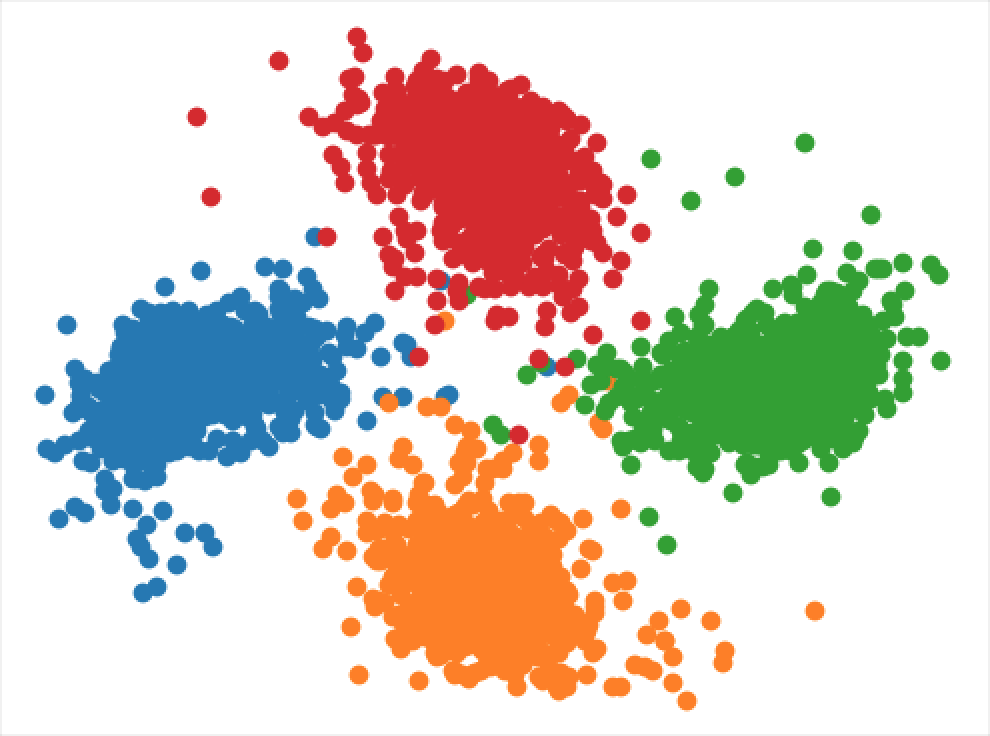}}
    \hfill
    \subcaptionbox{$\lambda=10.$}{\includegraphics[width=0.136\textwidth]{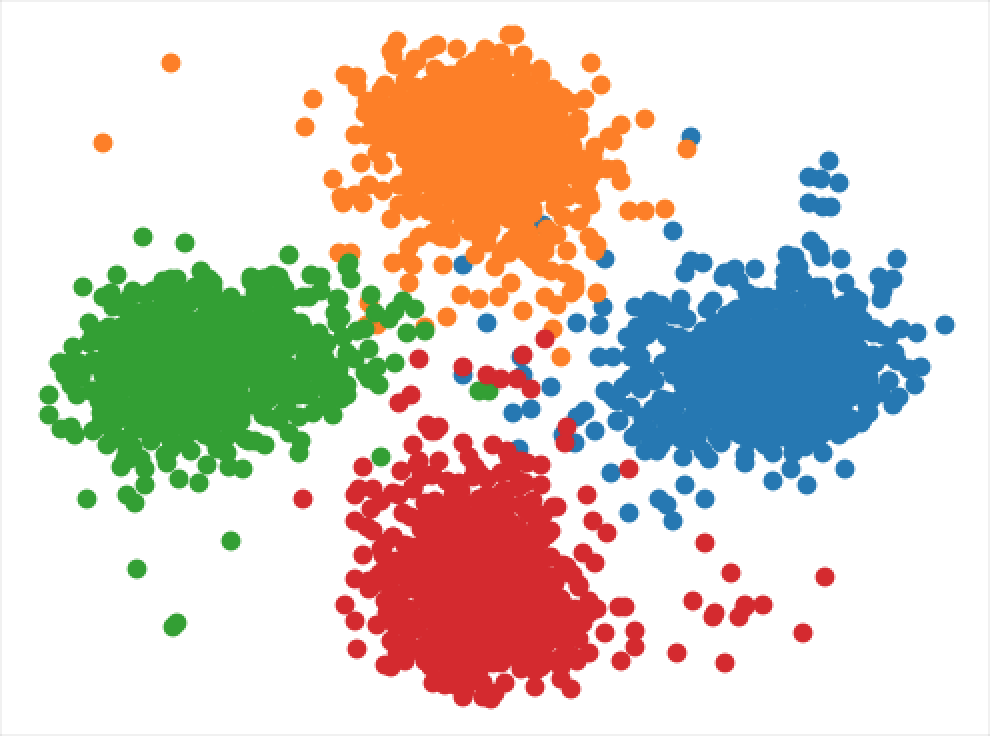}}
    \hfill
    \subcaptionbox{$\lambda=20.$}{\includegraphics[width=0.136\textwidth]{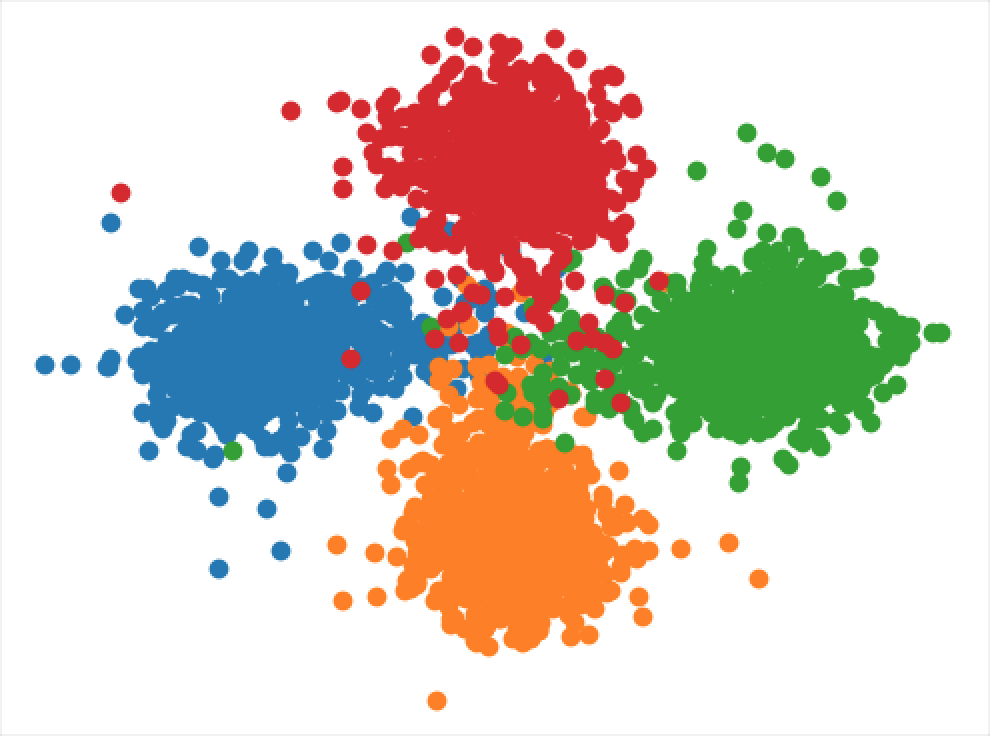}}\\
    \subcaptionbox{Data}{\includegraphics[width=0.136\textwidth]{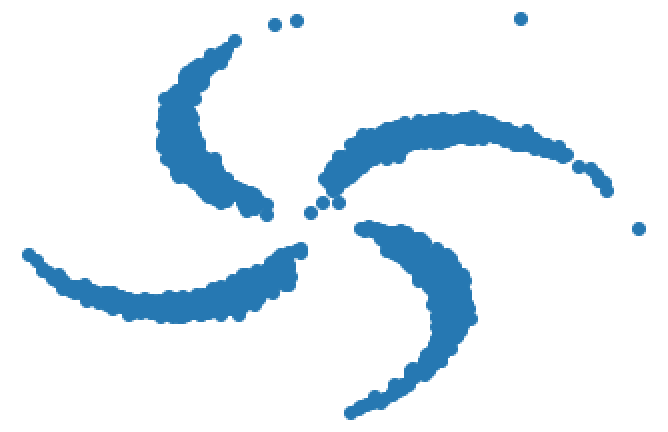}}
    \rulesep
    \hfill
    \subcaptionbox{$\lambda=0.5$}{\includegraphics[width=0.136\textwidth]{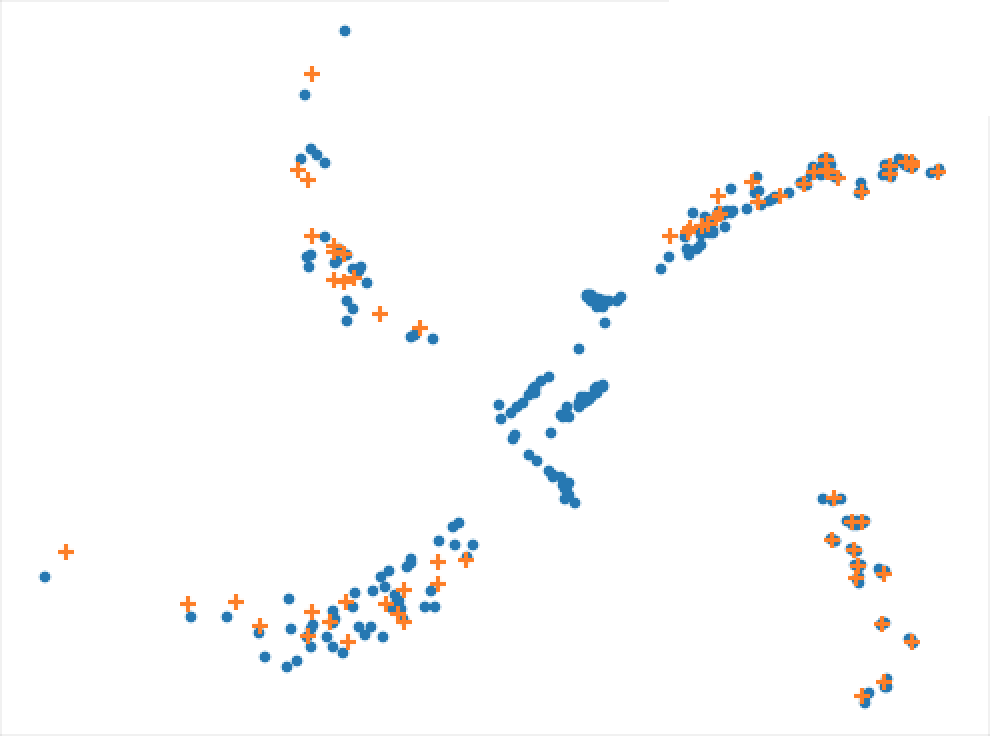}}
    \hfill
    \subcaptionbox{$\lambda=1.0$}{\includegraphics[width=0.136\textwidth]{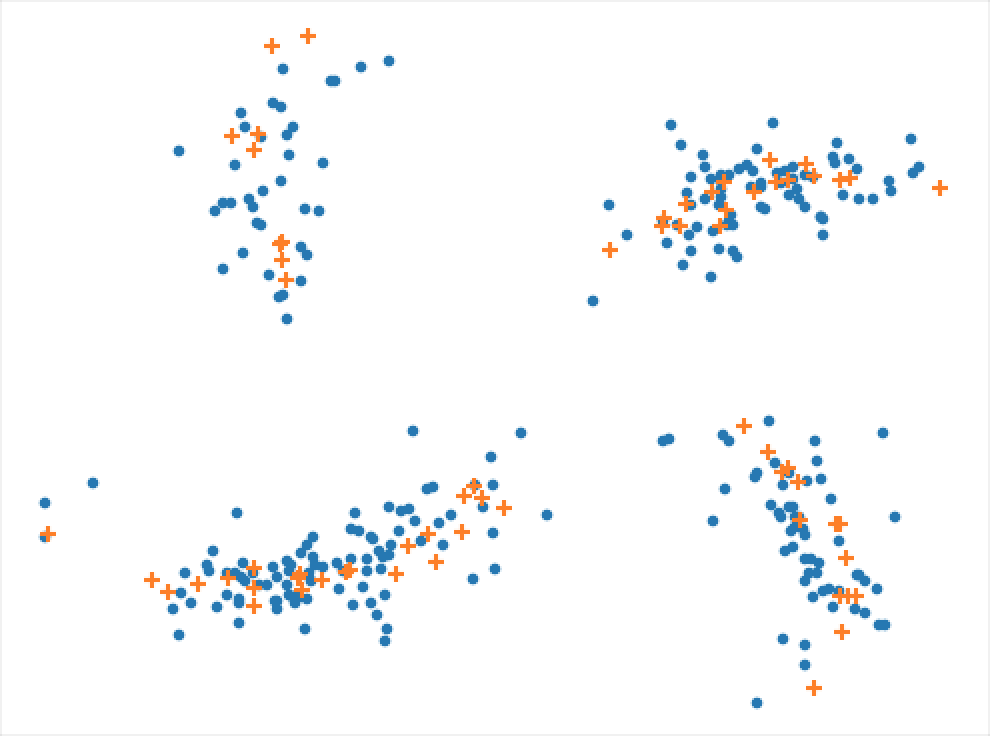}}
    \hfill
    \subcaptionbox{$\lambda=2.5$}{\includegraphics[width=0.136\textwidth]{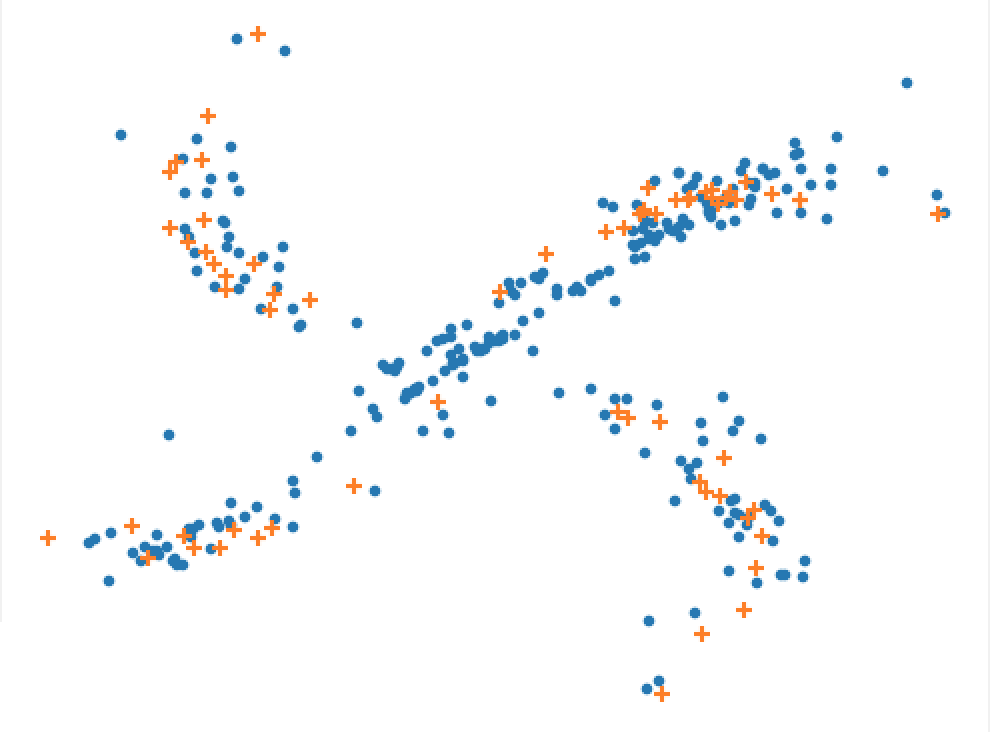}}
    \hfill
    \subcaptionbox{$\lambda=5.$}{\includegraphics[width=0.136\textwidth]{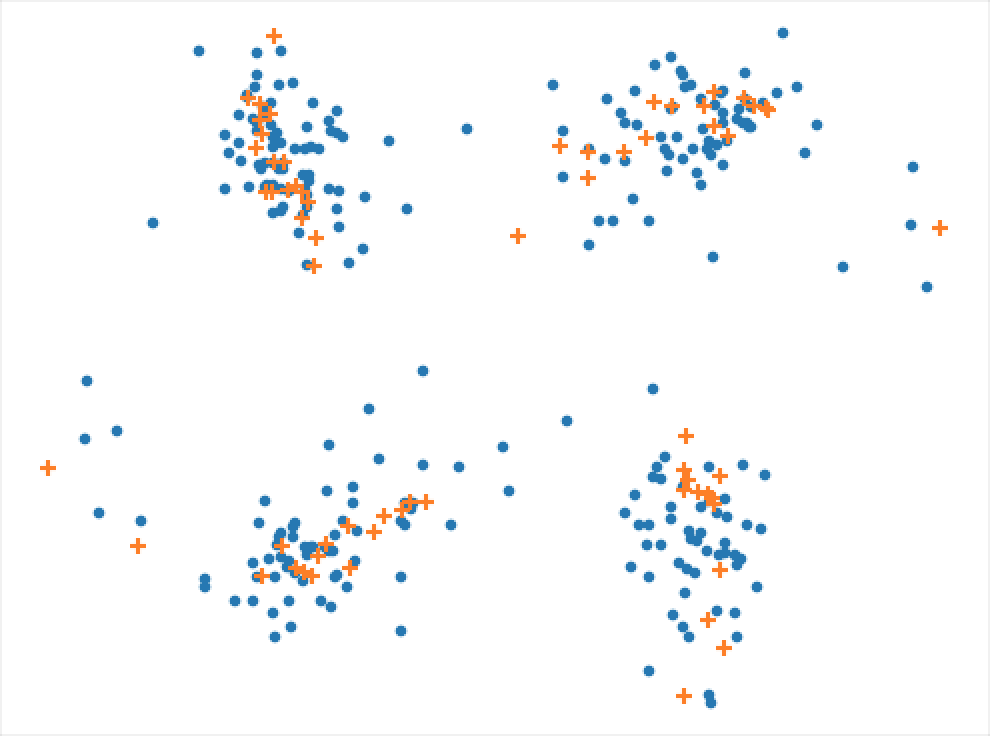}}
    \hfill
    \subcaptionbox{$\lambda=10.$}{\includegraphics[width=0.136\textwidth]{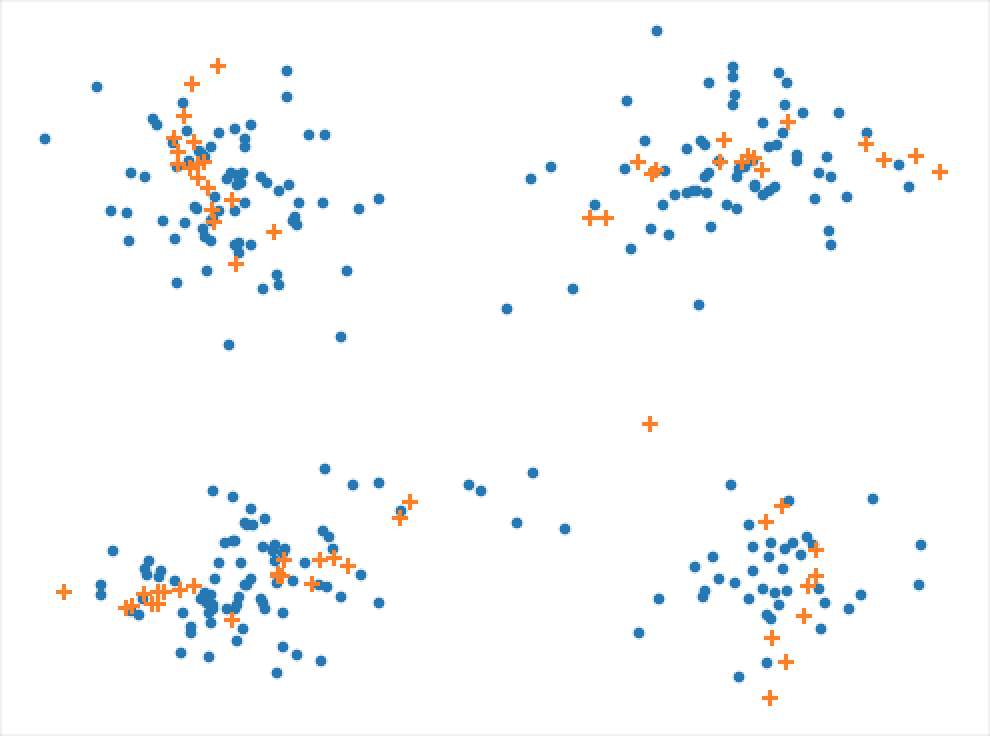}}
    \hfill
    \subcaptionbox{$\lambda=20.$}{\includegraphics[width=0.136\textwidth]{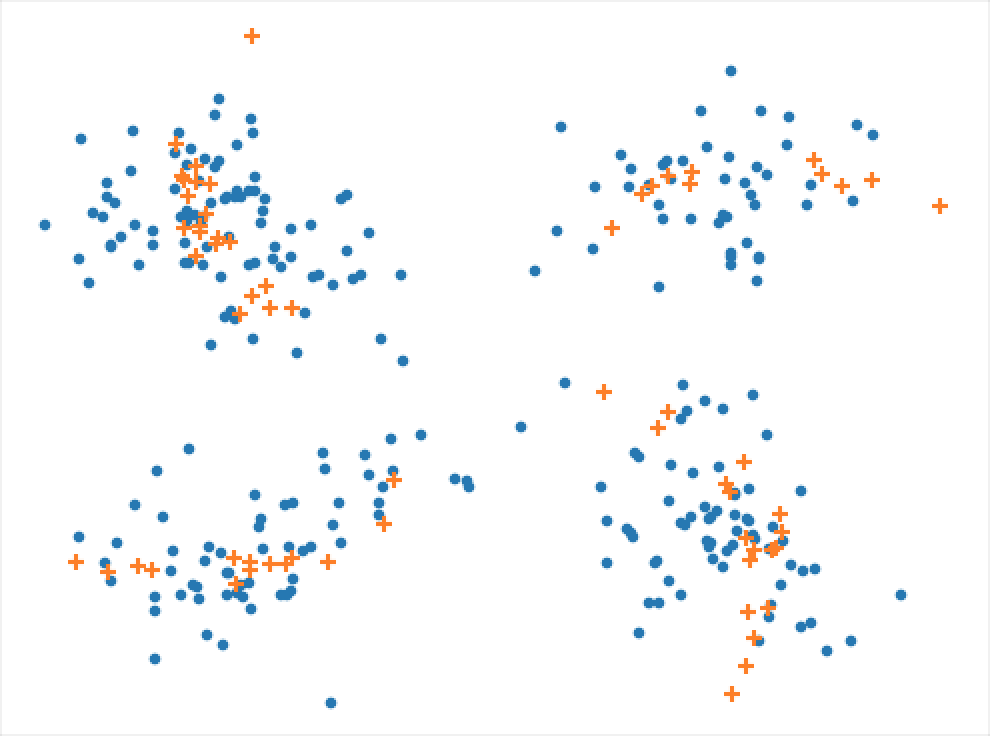}}
    \caption{Visualization of latent embeddings (row 1) and reconstruction (row 2) for different $\lambda$ for pinwheel dataset with mixture of Gaussian prior. The colors in the first rows b)-g) represents the true data class. Best viewed in color. }
    \label{fig:alpha_ablation_pinwheel}
\end{figure*}

\textbf{Datasets.} We consider two toy datasets used for Section~\ref{subsec:empirical_analysis}. A 1D mixture of two Gaussians $p(z) = \frac{1}{2} \mathcal{N}(z;-3, 1) + \frac{1}{2}\mathcal{N}(z; 3, 1)$ is used to visualize the challenge of fitting a univariate Gaussian to a Gaussian mixture. We use 2000 i.i.d. samples for training. Further, we
use the ``pinwheels'' dataset from~\cite{johnson2016composing}. We generated spiral cluster data with $N = 4000$ observations, equally clustered in four spirals with radial and tangential standard deviations respectively of $0.05$ and $0.25$, and a rate of $0.25$. For density estimation, kNN clustering, and semi-supervised learning we carried out experiments using five image datasets: static MNIST~\cite{larochelle2011neural}, dynamic MNIST~\cite{salakhutdinov2008quantitative}, Omniglot~\cite{lake2015human}, Caltech 101 Silhouette~\cite{marlin2010inductive} and CIFAR10~\cite{krizhevsky2009learning}.
For semi-supervised facial action unit recognition we use DISFA~\cite{mavadati2013disfa} and FERA2015~\cite{valstar2015fera}. For both datasets DISFA and FERA2015 the frames with intensities equal or greater than 2 are considered as positive while others are treated as negative. Further, we performed subject-independent 3-fold cross-validation for these two datasets. Details about all datasets can be found in the Appendix (3.1).

\textbf{Model architecture.} For the toy experiment with the Pinwheel dataset we used a neural network with two fully-connected and Softplus activation. In the experiments for MNIST (static and dynamic), Omniglot, and Caltech 101 Silhouettes we modeled all distributions using fully-connected neural networks with two hidden layers of $300$ hidden units in the unsupervised setting.
For CIFAR10 we employed a convolutional architecture with residual blocks similar to \cite{van2017neural} whereas for DISFA and FERA2015 we used a convolutional encoder and decoder pair. The detailed listing of all architecture used can be found in the Appendix (3.1).
For the toy dataset we used a Gaussian likelihood, for all colored images we used the discretized logistic likelihood as in~ \cite{kingma2016improved}, and for the other datasets we used a Bernoulli likelihood.

\textbf{Optimization and hyperparameters} 
All model weights of the neural networks were initialized according to~\cite{glorot2010understanding}.
For fitting a mixture of two Gaussians in Subsection~\ref{subsec:empirical_analysis}, we use gradient descent with a learning rate of $0.001$ for both KL and CS minimization. 
For training all other models, we used the ADAM algorithm~\cite{kinga2015method}, where we set the learning rate to $5 \cdot 10^{-4}$ and mini-batches of size $100$.
Additionally, we used the linear warm-up~\cite{bowman2015generating} for 100 epochs to avoid early collapse of the latent variable due to the divergence regularization.
During training, we used early-stopping with a look ahead of $100$ iterations to prevent over-fitting. 
For semi-supervised learning of DISFA and FERA2015 we perform optimization in two phases. First, we only train in an unsupervised fashion without any labels. Subsequently, we used the pre-trained model to include semi-supervised training with labels. Further, due to the imbalance of both datasets' label distribution, we also used iterative balanced batches during training. For further details, we refer to the Appendix (Section 3.2).

\textbf{Evaluation metrics} We report all evaluation metrics on the test set based on the best validation loss from training. In fact, the log-marginal likelihood (LL) has been the default evaluation metric for models optimizing the LL or ELBO.
The marginal likelihood can be computed generating $S$ samples from the recognition model using importance sampling and using the following estimator:
\begin{align}
p(x) \approx \frac{1}{S} \sum_{s=1}^S \frac{p(x|f(\xi^{(s)}))p(\xi^{(s)})}{q(\xi^{(s)|x)})}, \; \xi^{(s)} \sim q(\xi|x).
\end{align}
However, Theis et al.~\cite{theis2016note} showed that high LL does not necessarily correspond to plausible samples and thus is not a suitable metric for assessing image quality. Furthermore, in our own experiments, we observed that while the LL often increased by a large margin, neither the FID nor manual inspection showed improvement in image quality. For these reasons we report the \fid{ } (FID) for density estimation. The FID is a measure of similarity between two datasets of images and is often used to evaluate the fidelity of samples from Generative Adversarial Networks~\cite{goodfellow2014generative}. Heusel et al.~\cite{heusel2017gans} showed that this measure correlates with human perception of visual quality and can detect mode collapses in contrast to Inception Score (IS)~\cite{SalimansGZCRCC16}.

The \fid uses embeddings from the Inception v3 model to calculate the means and covariances of the real samples $(m, C)$ and the generated samples $(m_w, C_w)$. The metric is calculated using the Wasserstein-2 distance between these means and covariances.
\begin{align}
\begin{split}
   d^2((m, C),(m_w, C_w)) &=  \norm{m-m_w}^2_2  \\ &\quad + \textrm{Tr}(C+C_w-2(CC_w)^{1/2}).
\end{split}
\end{align}

Further, for clustering classification error rate and for facial action unit recognition F1 score was used for evaluation.
\subsection{Empirical analysis}\label{subsec:empirical_analysis}
\begin{figure*}[t]
    \centering
    \subcaptionbox{Trade-off between test reconstruction (RE) and test CS divergence with respect to $\lambda$. The scatter marker size indicate the value of $\lambda$ (the greater the marker size the higher $\lambda$).\label{subfig:lambda-cs-recon} }{\includegraphics[width=0.3\textwidth]{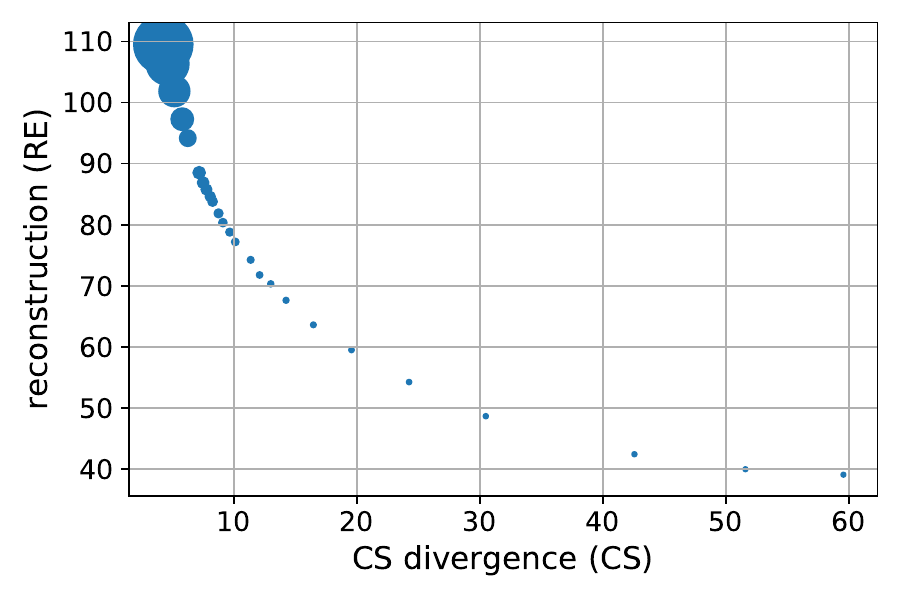}}
    \hfill
    \subcaptionbox{Relationship between model selection criterion $\min (\E_q(\log p(x|z)) - \textrm{D}_{\textrm{CS}}(q \parallel p))$ (RE + CS, lower is better) and \fid (FID, lower is better)\label{subfig:recs-fid}}{\includegraphics[width=0.3\textwidth]{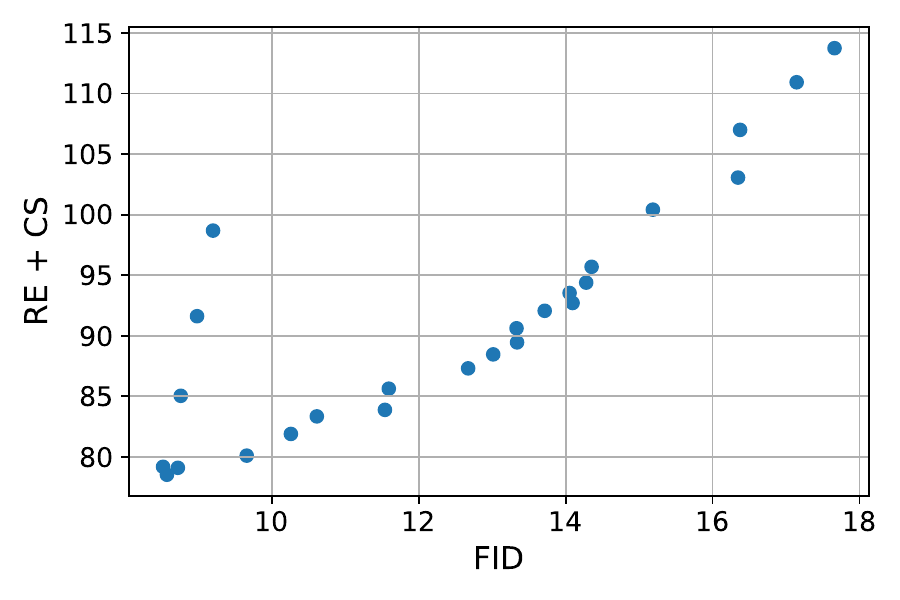}}
    \hfill
    \subcaptionbox{Relationship between model selection criterion $\min (\E_q(\log p(x|z)) - \textrm{D}_{\textrm{CS}}(q \parallel p))$ (RE + CS, lower is better) and log-marginal likelihood (LL, lower is better)\label{subfig:recs-ll}}{\includegraphics[width=0.3\textwidth]{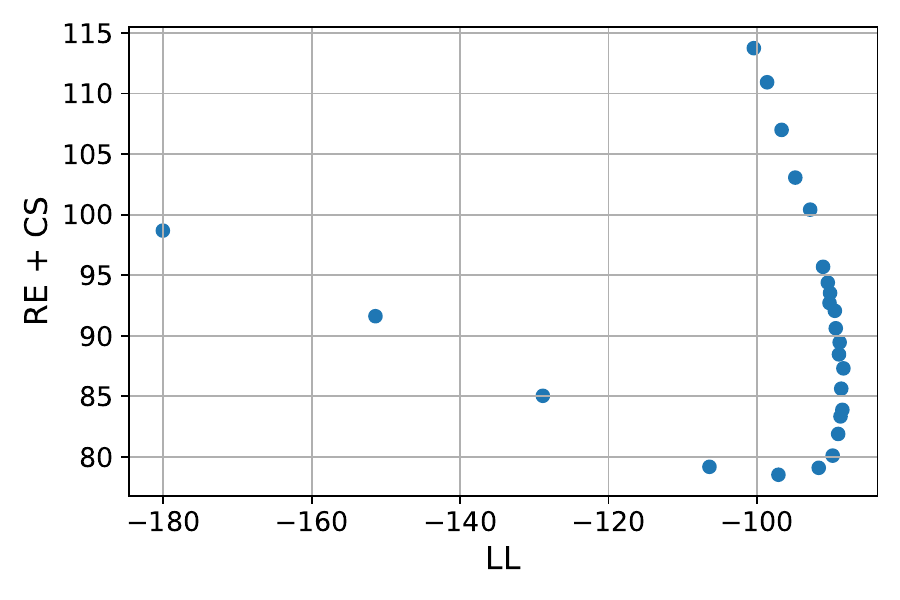}}
    \caption{Ablation for demonstrating effect of constraint $\lambda$ on model selection criterion, log-marginal likelihood (LL) and \fid (FID). All experiments were conducted with dynamic MNIST. We considered $\lambda$ in range of [0.25, 0.5, 0.75, \ldots, 9.5, 9.75, 10.0].}\label{fig:ablation}
\end{figure*}
\textbf{Fitting a mixture of two Gaussians}
Consider a one-dimensional mixture of Gaussians as the posterior of interest
\begin{align}
p(z) = \frac{1}{2} \mathcal{N}(z; -3, 1) + \frac{1}{2} \mathcal{N}(z; 3, 1).
\end{align}
The posterior contains multiple modes. We seek to approximate it with two objectives: Kullback-Leibler (KL) with a Gaussian approximating family and Cauchy-Schwarz (CS) with a mixture of Gaussians approximating family. In both settings we can calculate the divergence analytically. Figure~\ref{fig:toy_mixture}(a) displays the posterior approximations. We find that the KL divergence chooses a single mode and has slightly different variances. This approximation does not produce good results because a single Gaussian is a poor approximation of the mixture. The approximate posterior in Figure~\ref{fig:toy_mixture} (b) comes from a using a mixture of Gaussian prior. Because this enables the prior to be more general it can capture the posterior.

\textbf{Visualizing the effect of $\lambda$.} We made the assumption that constrained optimization is important for enabling CSRAE models to learn multi-modal representations. One way to view $\lambda$ is as a coefficient balancing the  reconstruction and and prior-matching term of the CSRAE object. We visualized this effect in Figure~\ref{fig:alpha_ablation_pinwheel} where we applied our MixtureCSRAE objective from~\eqref{eq:mixturecsrae} to the pinwheel toy dataset. The prior visualized in Figure~\ref{fig:alpha_ablation_pinwheel} (a) is defined as

\begin{align}
    p(z) = &\; \sum_{k=1}^{K} \pi^k \mathcal{N}(z; \mu_{k}, \sigma_{k})\\
    = &\; \sum_{k=1}^{K} \pi^k  \prod_{d=1}^D \mathcal{N}(z_d, \mu^{k}_{d}, \sigma^{k}_{d}),
\end{align}
with $D=2$, $K=4$, $\sigma^k = 0.05 I_D$, $\pi^k = \frac{1}{K}$ and $\mu_d^k \in \{0, 1\}$.
We observe that with a low $\lambda$ value, e.g., $\lambda=0.5$, the model can reconstructthe input data (Figure~\ref{fig:alpha_ablation_pinwheel} (i)) almost perfectly, however, the structure of the latent variable depicted in Figure~\ref{fig:alpha_ablation_pinwheel} (b) is not similar to prior visualized in Figure~\ref{fig:alpha_ablation_pinwheel} (a). As $\lambda$ is increased, the latent space embedding (Figures~\ref{fig:alpha_ablation_pinwheel} (b-g)) draws nearer to the prior, the reconstruction decreases in quality resulting the reconstructed datapoints to be less aligned to the original input data.

\textbf{Model selection and relationship to LL and FID.} As shown in Figure~\ref{fig:alpha_ablation_pinwheel} there is a trade-off between the reconstruction and prior-matching-term--the two terms which make up our loss objective. The trade-off is influenced by $\lambda$. The higher $\lambda$ the more weight is put on the approximate posterior matching the prior and the less weight is put on the reconstruction error. We conducted ablation studies with CSRAE on dynamic MNIST shown in Figure~\ref{fig:ablation} to understand the effect of $\lambda$ on the loss terms and the relationship between loss objective and evaluation metrics FID and LL. Figure~\ref{subfig:lambda-cs-recon} shows reconstruction loss (y-axis) and prior-matching-term (x-axis) with respect to $\lambda$. In this Figure the marker size is an indicator of the value of $\lambda$, i.e., the greater the marker size the higher the value of $\lambda$. We can observe a gradual decrease in reconstruction error with decreasing $\lambda$ while simultaneously the CS term increases. We added both terms together and plotting them against LL (Figure~\ref{subfig:recs-ll}) and FID (Figure~\ref{subfig:recs-fid}). The plots visualizes that our model selection criteria $\min \Big(\E_{q_{\phi}}[\log p_{\theta}(x|z)] + \textrm{D}_{\textrm{CS}}(q_{\phi}(z|x) \parallel p(z))\Big)$ seems proportional to evaluation metrics LL and FID. When our model selection criteria is lowest, FID is at its lowest as well. For LL, we also observe a low $\min \Big(\E_{q_{\phi}}[\log p_{\theta}(x|z)] + \textrm{D}_{\textrm{CS}}(q_{\phi}(z|x) \parallel p(z))\Big)$ means low LL, however, LL does not necessarily obtain its minimum at the lowest model selection criteria.

\begin{figure}[t]
    \centering
    \includegraphics[width=0.6\linewidth]{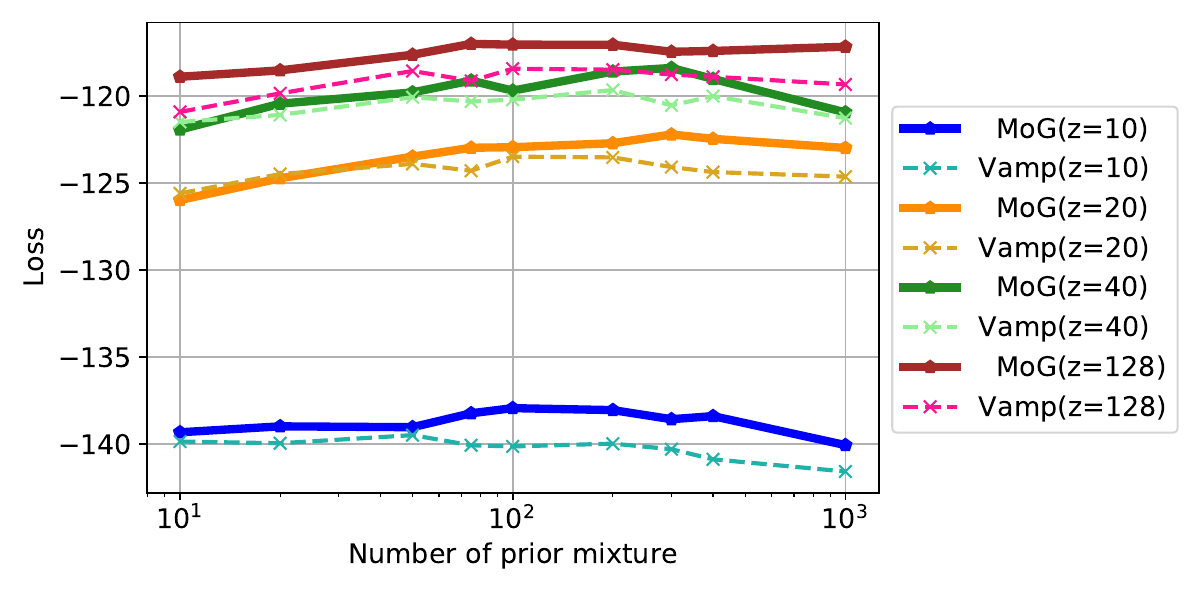}
    \caption{Comparing MoGPrior and VampPrior with varying latent variable dimensions and number of mixtures. We report average ($n=5$) loss (greater is better), c.f. model selection criterion. All solid lines represent MoGPrior with different number of latent dimensions, while all dashed lines represent VampPrior. }
    \label{fig:ablation:mixture}
\end{figure}

\textbf{Number of mixture components.}
We compared our MixtureCSRAE with varying number of mixture components. Figure~\ref{fig:ablation:mixture} visualizes the LL depending on the number of mixture components ($10, 20, 40, 100, 200, 300, 400, 1000$) and the number of latent dimensions ($10, 20, 40, 128$). We observe a trend of increased performance with increasing number of mixture components. But this trend is not retained for larger number of mixtures ($> 400$) as performance either drops or remain unchanged. For MoGPrior this drop in performance could be due to increased difficulty to learn a large GMM prior while simultaneously optimizing the approximate posterior parameters. For VampPrior an increase in number of mixture components could aggravate overfitting and thus lead to a performance decrease.

\textbf{MoGPrior vs. VampPrior.} When comparing MoGPrior and VampPrior in our ablation study visualized in Figure~\ref{fig:ablation:mixture}, we can see that the learned prior (MoGPrior) either is of similar or superior performance to VampPrior. The difference in performance is usually smaller for lower number of mixture components ($<100$). However, the gap is more obvious for larger number of mixture components and larger latent dimensions. As mentioned before overfitting of VampPrior might be the reason for this gap.
\subsection{Density estimation on common benchmarks}\label{subsec:density}
\begin{table}[t]
    \centering
        \begin{tabular}{l c c c c c}
            \toprule
            \multirow{3}{*}{\textbf{Model}} & \multicolumn{5}{c}{\textbf{Dataset}} \\
            \cmidrule{2-6}
            & Static & Dynamic & \multirow{2}{*}{Omniglot} & \multirow{2}{*}{Caltech101} & \multirow{2}{*}{CIFAR10}\\ 
            & MNIST & MNIST & & &\\
            \toprule
            VAE~\cite{kingma2013auto} ($z=40$) &
                10.04 \pms{0.01} &
                10.99 \pms{0.01} & 
                9.09 \pms{0.01} &
                18.08 \pms{0.02} &
                11.36 \pms{0.01} \\
            \midrule
            VAE~\cite{kingma2013auto} ($z=128$) &
                10.23 \pms{0.02} &
                11.11 \pms{0.01} &
                8.95 \pms{0.02} &
                17.89 \pms{0.04} &
                11.58 \pms{0.01} \\
            \midrule
            IWAE~\cite{burda2015importance} &
                \multirow{2}{*}{9.87 \pms{0.01}} &
                \multirow{2}{*}{10.97 \pms{0.01}} &
                \multirow{2}{*}{9.13 \pms{0.02}} &
                \multirow{2}{*}{18.03 \pms{0.02}} &
                \multirow{2}{*}{16.15 \pms{0.00}} \\
            ($z=40,n_{\textrm{iw}}=5$) & & & & & \\
            \midrule
            IWAE~\cite{burda2015importance} &
                \multirow{2}{*}{9.83 \pms{0.01}} &
                \multirow{2}{*}{10.92 \pms{0.01}} &
                \multirow{2}{*}{8.40 \pms{0.01}} &
                \multirow{2}{*}{17.47 \pms{0.03}} &
                \multirow{2}{*}{-} \\
            ($z=128,n_{\textrm{iw}}=50$) & & & & & \\
            \midrule
            VampPriorVAE~\cite{tomczak2018vae} &
                \multirow{2}{*}{11.64 \pms{0.03}} &
                \multirow{2}{*}{12.69 \pms{0.03}} &
                \multirow{2}{*}{8.88 \pms{0.01}} &
                \multirow{2}{*}{17.45 \pms{0.04}} &
                \multirow{2}{*}{11.19 \pms{0.02}} \\
            ($z=40,k=10$) & & & & & \\ 
            \midrule
            VampPriorVAE~\cite{tomczak2018vae} &
                \multirow{2}{*}{10.35 \pms{0.03}} &
                \multirow{2}{*}{11.35 \pms{0.04}} &
                \multirow{2}{*}{8.86 \pms{0.01}} &
                \multirow{2}{*}{17.70 \pms{0.04}} &
                \multirow{2}{*}{11.12 \pms{0.02}} \\
            ($z=40,k=100$) & & & & & \\
            \midrule
            VampPriorVAE~\cite{tomczak2018vae} &
                \multirow{2}{*}{10.01 \pms{0.04}} &
                \multirow{2}{*}{11.10 \pms{0.02}} &
                \multirow{2}{*}{8.76 \pms{0.01}} &
                \multirow{2}{*}{17.70 \pms{0.05}} &
                \multirow{2}{*}{11.09 \pms{0.03}} \\
            ($z=40,k=400$) & & & & & \\
            \toprule
            CSRAE ($z=40$) &
                \normalsize{\textbf{7.60}} \pms{0.01} &
                7.88 \pms{0.02} &
                8.13 \pms{0.02} &
                17.15 \pms{0.05} &
                10.94 \pms{0.03} \\
            \midrule
            MixtureCSRAE &
                \multirow{2}{*}{7.67 \pms{0.02}} &
                \multirow{2}{*}{\normalsize{\textbf{7.81}} \pms{0.02}} &
                \multirow{2}{*}{8.12 \pms{0.01}} &
                \multirow{2}{*}{17.44 \pms{0.03}} &
                \multirow{2}{*}{10.92 \pms{0.02}} \\
            ($z=40,k=10$) & & & & & \\ \hline
            MixtureCSRAE &
                \multirow{2}{*}{7.68 \pms{0.02}} &
                \multirow{2}{*}{7.89 \pms{0.01}} &
                \multirow{2}{*}{7.96 \pms{0.01}} &
                \multirow{2}{*}{\normalsize{\textbf{16.33}} \pms{0.03}} &
                \multirow{2}{*}{10.65 \pms{0.03}} \\
            ($z=40,k=100$) & & & & & \\ \hline
            MixtureCSRAE &
                \multirow{2}{*}{7.74 \pms{0.01}} &
                \multirow{2}{*}{8.03 \pms{0.01}} &
                \multirow{2}{*}{\normalsize{\textbf{7.90}} \pms{0.01}} &
                \multirow{2}{*}{16.66 \pms{0.03}} &
                \multirow{2}{*}{\normalsize{\textbf{10.54}} \pms{0.02}} \\
            ($z=40,k=400$) & & & & & \\
            \bottomrule
        \end{tabular}
    \caption{Average test \fid{} (FID, $n=5$) and standard error (lower is better). For CIFAR10, IWAE~\cite{burda2015importance} was not computed for dimensions $z=128$ and number of importance weights $k=50$, as this would have been too computationally expensive due to the residual networks used for encoder and decoder.}\label{tab:density}
\end{table}

\textbf{Quantitative results.} We quantitatively evaluate our method using the FID. In Table~\ref{tab:density} we present a comparison between our proposed approach (CSRAE, MixtureCSRAE) and variational auto-encoding models VAE \cite{kingma2013auto}, IWAE \cite{burda2015importance} and VampPriorVAE \cite{tomczak2018vae}. The comparison includes training MLP-based models for static and dynamic MNIST, Omniglot and Caltech 101 Silhouettes as well convolutional model with residual blocks for CIFAR10. For fair comparison, we trained all models with the same optimization scheme and model architecture. 
First, we notice that in all cases the application of the CSRAE and MixtureCSRAE results in a substantial improvement of the generative performance in terms of the test FID, accounting for at least 4\% (CIFAR10) and up to 22\% (Dynamic MNIST) improvement in performance. 
Further, for more complex datasets like Omniglot, Caltech101, and CIFAR10, we also observe that a multi-modal prior (MixtureCSRAE) improves upon performance compared to a simple Gaussian prior (CSRAE).

\begin{figure*}[ht!]
    \centering
    \subcaptionbox{Static MNIST (IWAE)}{\includegraphics[width=0.32\textwidth]{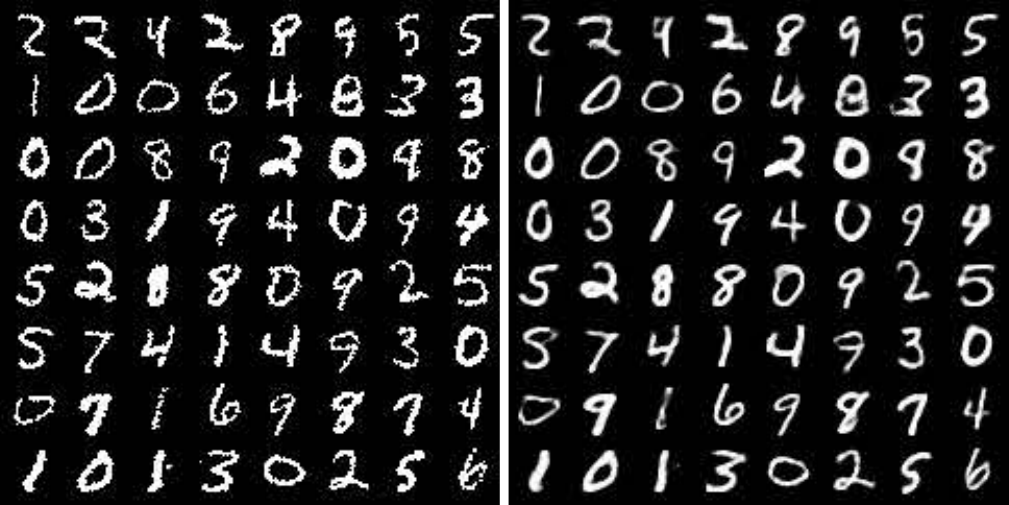}}
    \subcaptionbox{Static MNIST (VampPriorVAE)}{\includegraphics[width=0.32\textwidth]{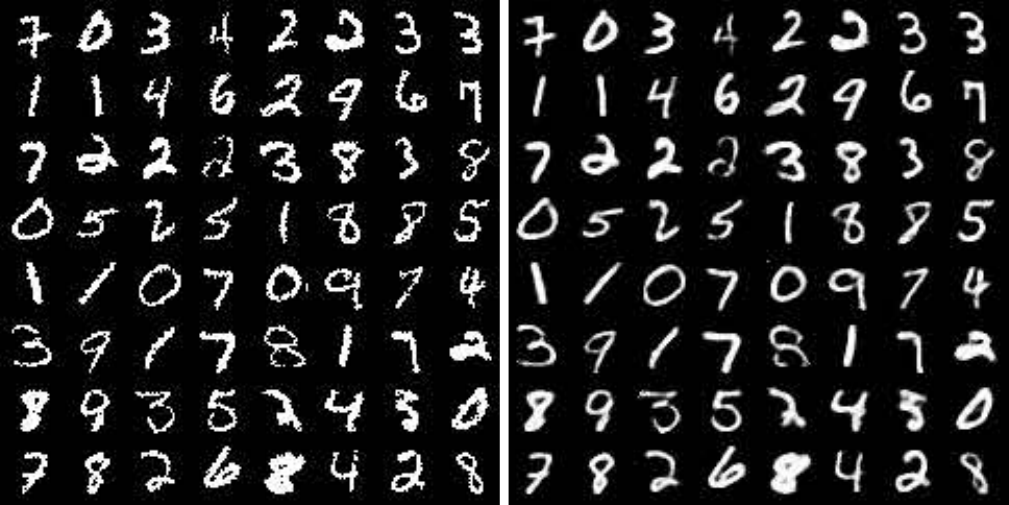}}
    \rulesep \rulesep
    \subcaptionbox{Static MNIST (MixtureCSRAE)}{\includegraphics[width=0.32\textwidth]{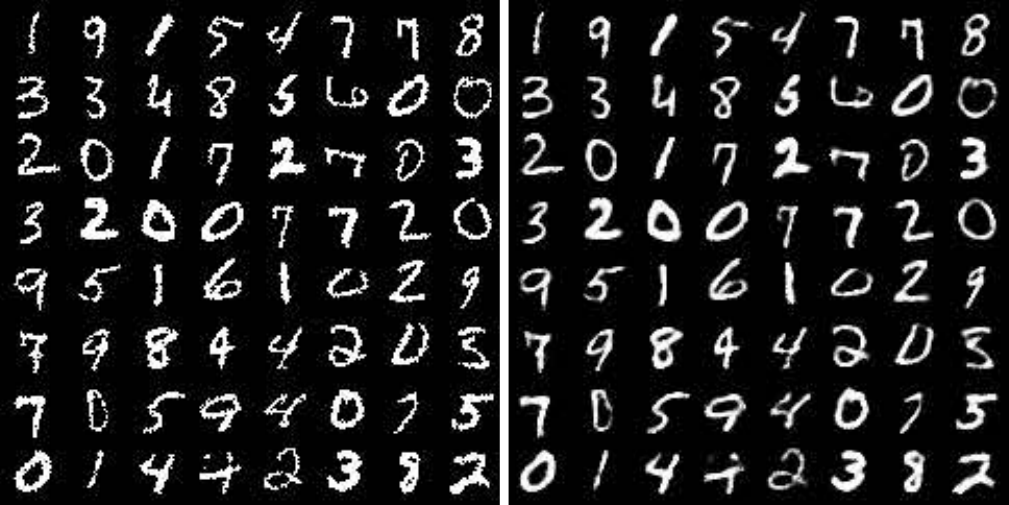}}
    \subcaptionbox{Dynamic MNIST (IWAE)}{\includegraphics[width=0.32\textwidth]{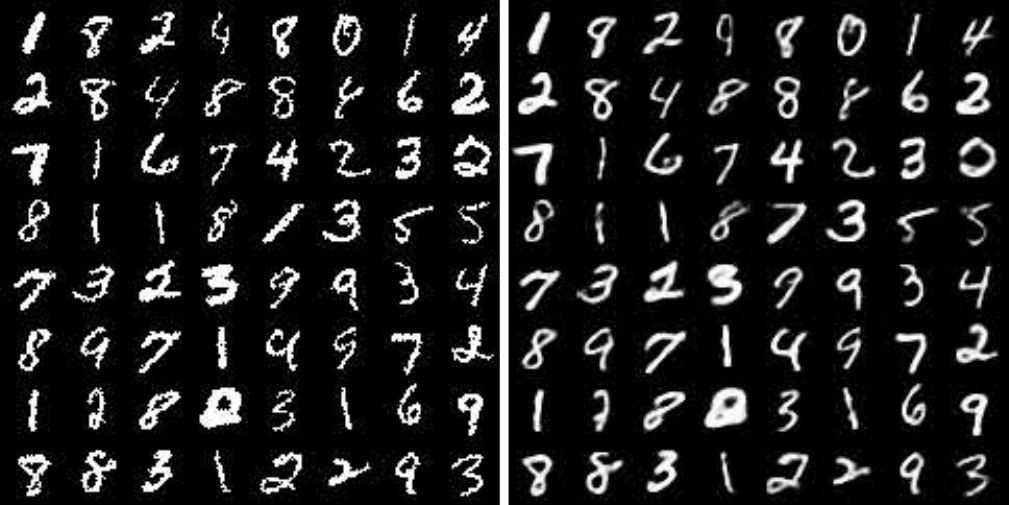}}
    \subcaptionbox{Dynamic MNIST  (VampPriorVAE)}{\includegraphics[width=0.32\textwidth]{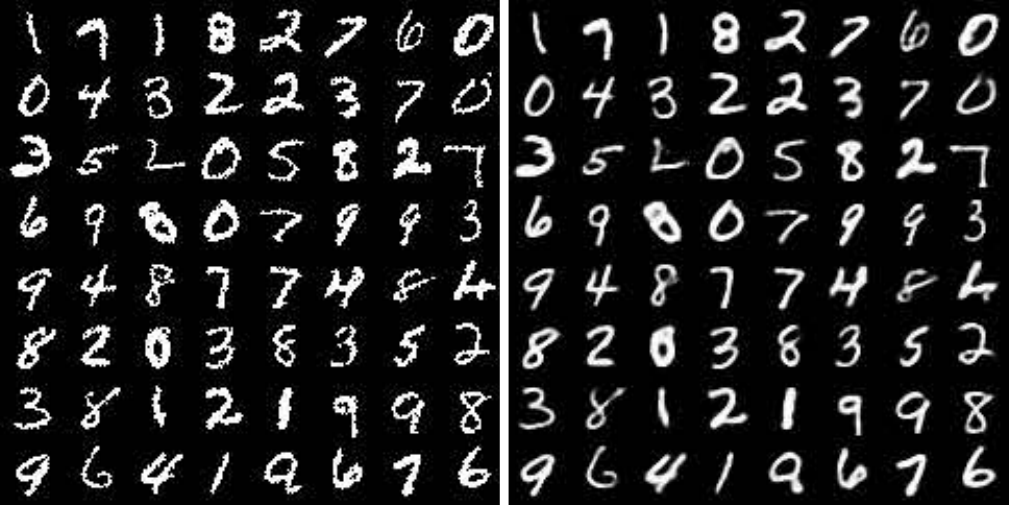}}
    \rulesep \rulesep
    \subcaptionbox{Dynamic MNIST  (MixtureCSRAE)}{\includegraphics[width=0.32\textwidth]{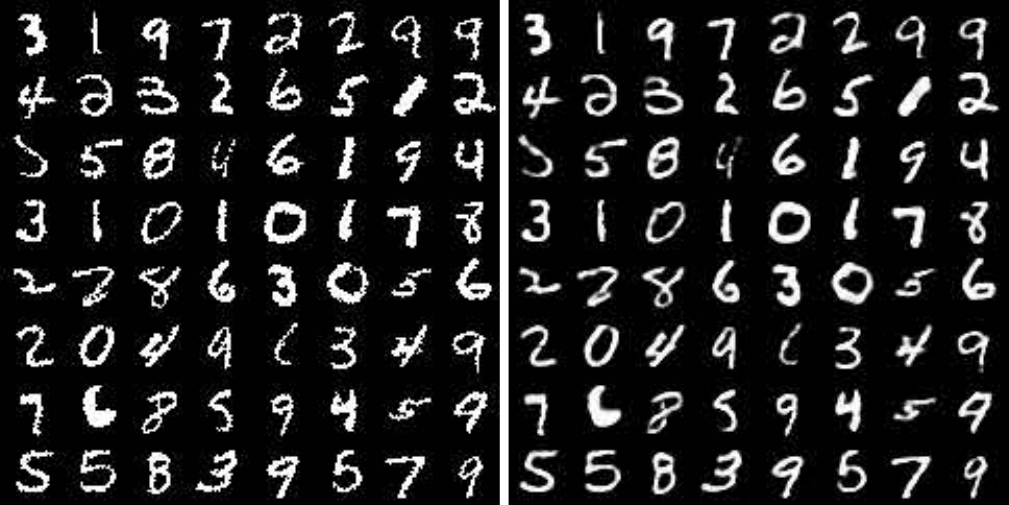}}
    \subcaptionbox{Omniglot (IWAE)}{\includegraphics[width=0.32\textwidth]{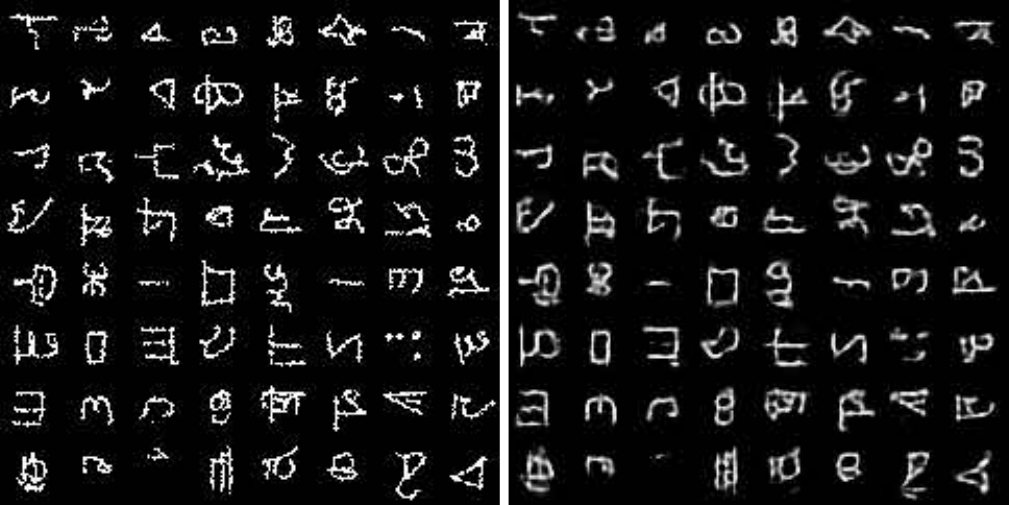}}
    \subcaptionbox{Omniglot  (VampPriorVAE)}{\includegraphics[width=0.32\textwidth]{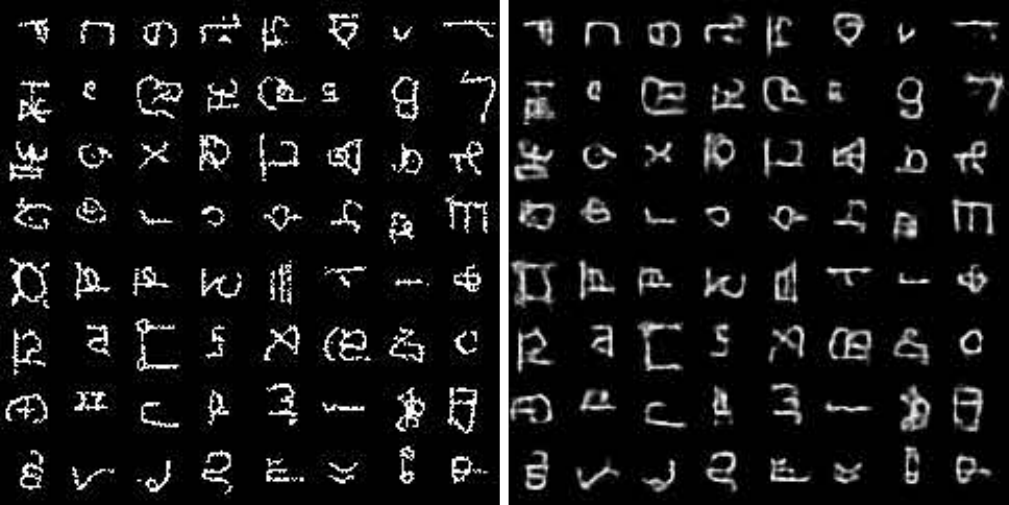}}
    \rulesep \rulesep
    \subcaptionbox{Omniglot  (MixtureCSRAE)}{\includegraphics[width=0.32\textwidth]{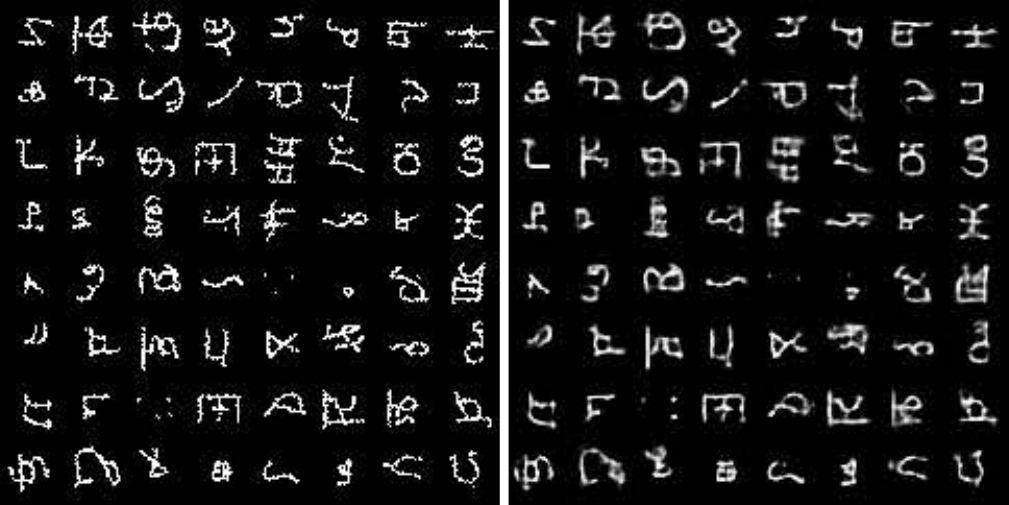}}
    \subcaptionbox{Caltech101  (IWAE)}{\includegraphics[width=0.32\textwidth]{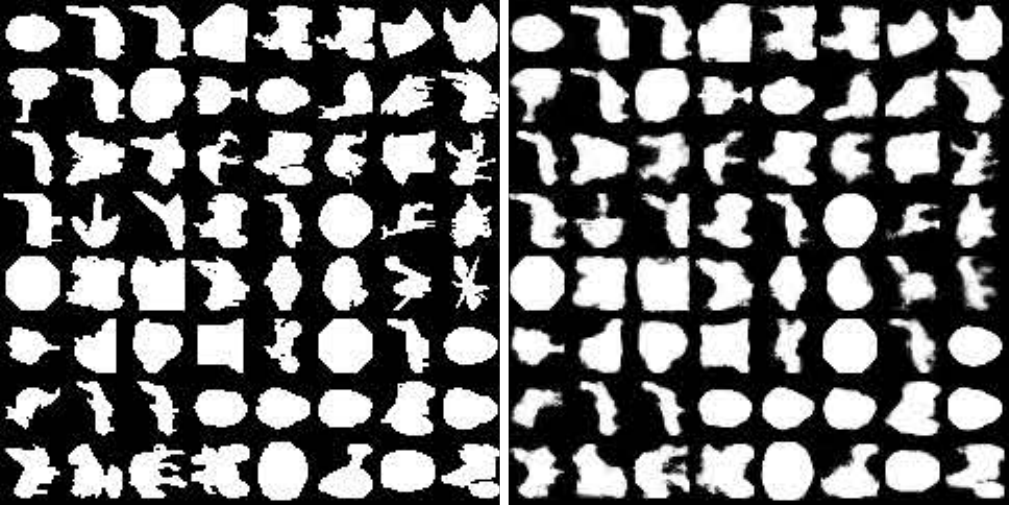}}
    \subcaptionbox{Caltech101  (VampPriorVAE)}{\includegraphics[width=0.32\textwidth]{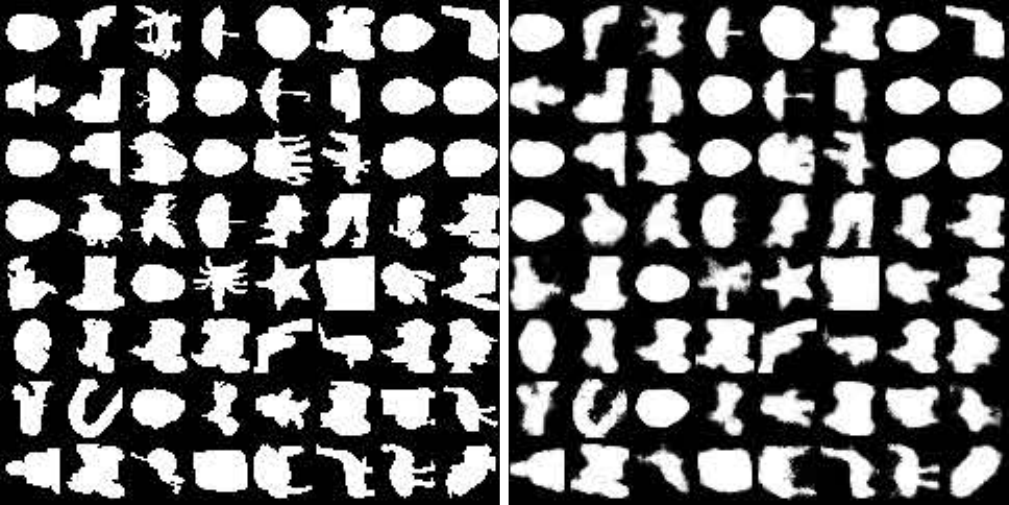}}
    \rulesep \rulesep
    \subcaptionbox{Caltech101 (MixtureCSRAE)}{\includegraphics[width=0.32\textwidth]{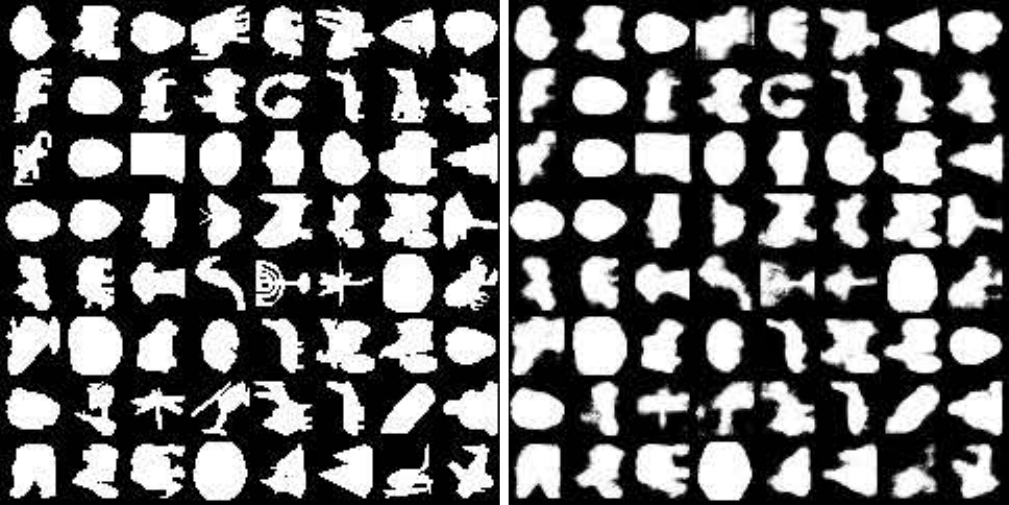}}
    \subcaptionbox{CIFAR10  (IWAE)}{\includegraphics[width=0.32\textwidth]{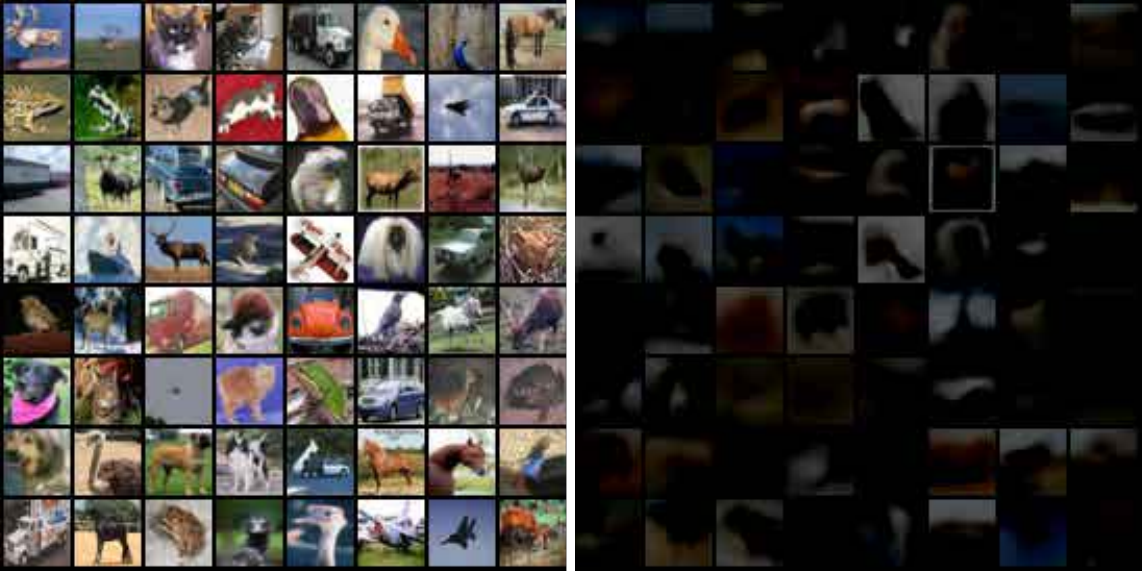}}
    \subcaptionbox{CIFAR10  (VampPriorVAE)}{\includegraphics[width=0.32\textwidth]{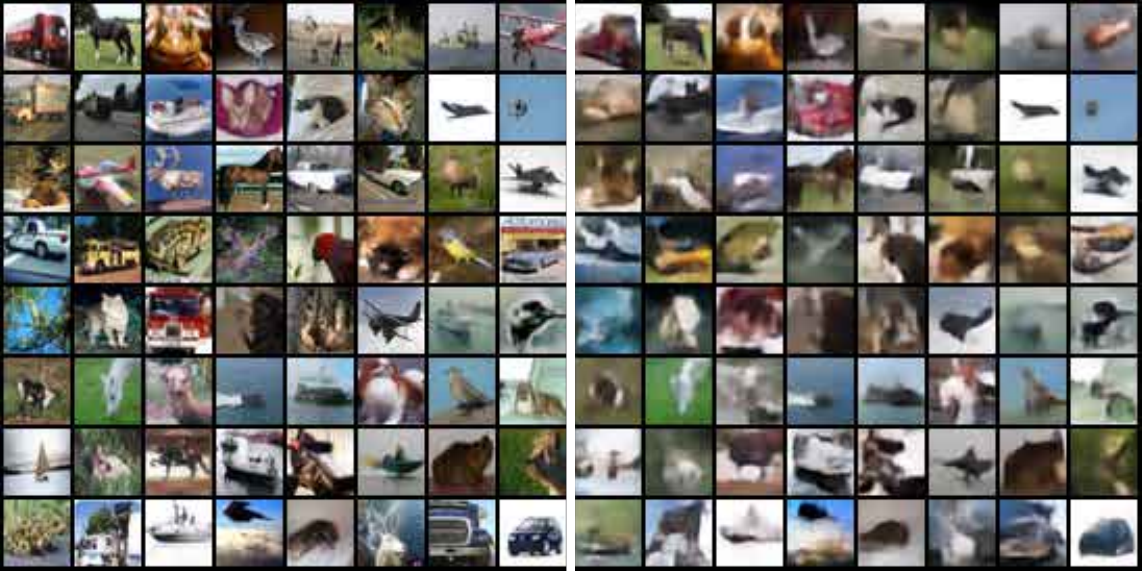}}
    \rulesep \rulesep
    \subcaptionbox{CIFAR10 (MixtureCSRAE)}{\includegraphics[width=0.32\textwidth]{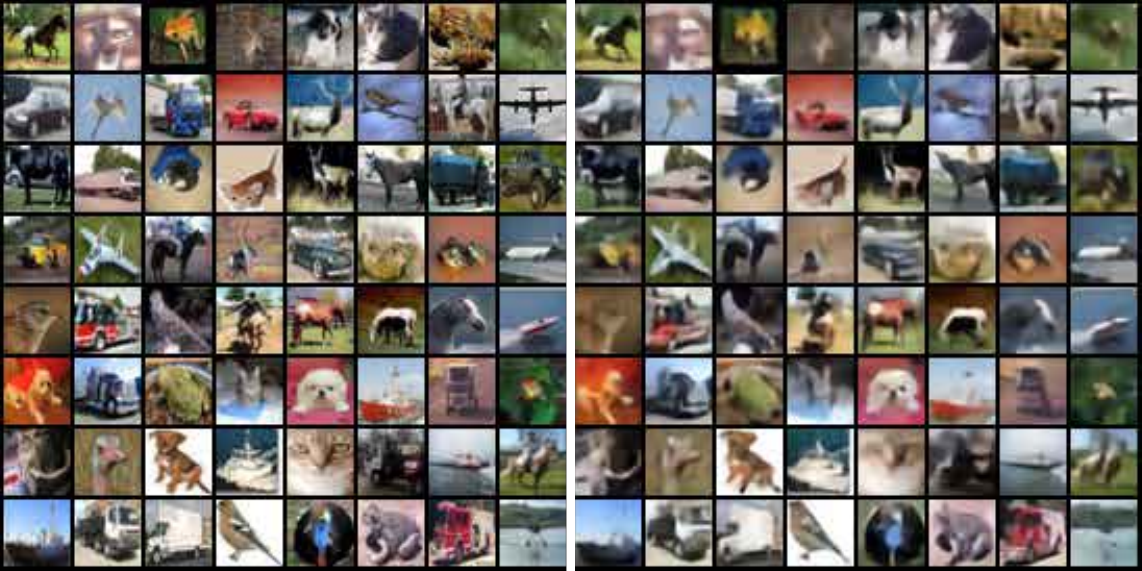}}
    \caption{Test samples and reconstructions of Static MNIST (a)-(c), Dynamic MNIST (d)-(f), Omniglot (g)-(i), Caltech101 (j)-(k) and CIFAR10 (m-o). We showed samples and reconstruction of IWAE (first column), VampPriorVAE (second column) and MixtureCSRAE (third column).}\label{fig:qual:rec}
    \hfill
\end{figure*}

\begin{figure*}[ht!]
    \centering
    \subcaptionbox{Static MNIST }{\includegraphics[width=0.175\textwidth]{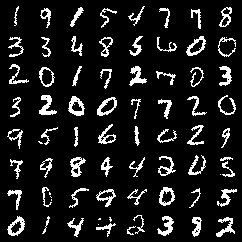}}
    \subcaptionbox{IWAE}{\includegraphics[width=0.175\textwidth]{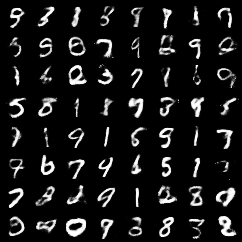}}
    \subcaptionbox{VampPriorVAE}{\includegraphics[width=0.175\textwidth]{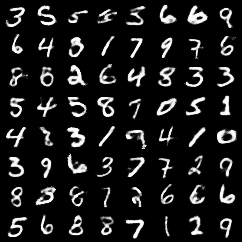}}
    \rulesep \rulesep
    \subcaptionbox{MixtureCSRAE}{\includegraphics[width=0.175\textwidth]{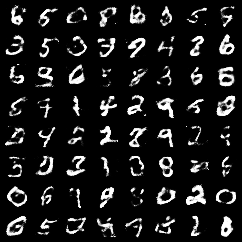}}
    \hfill \\
    \subcaptionbox{Dynamic MNIST}{\includegraphics[width=0.175\textwidth]{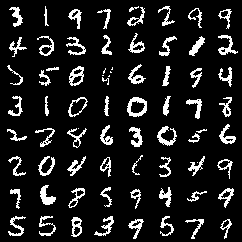}}
    \subcaptionbox{IWAE}{\includegraphics[width=0.175\textwidth]{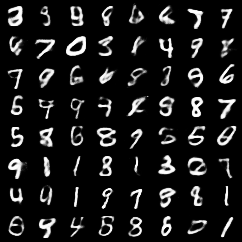}}
    \subcaptionbox{VampPriorVAE}{\includegraphics[width=0.175\textwidth]{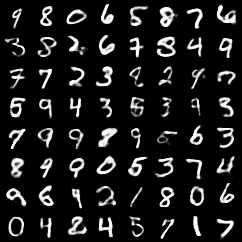}}
    \rulesep \rulesep
    \subcaptionbox{MixtureCSRAE}{\includegraphics[width=0.175\textwidth]{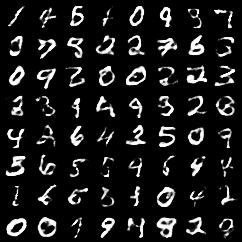}}
    \hfill \\
    \subcaptionbox{Omniglot }{\includegraphics[width=0.175\textwidth]{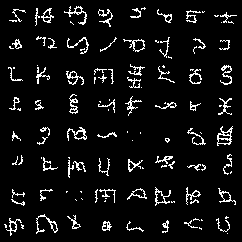}}
    \subcaptionbox{IWAE}{\includegraphics[width=0.175\textwidth]{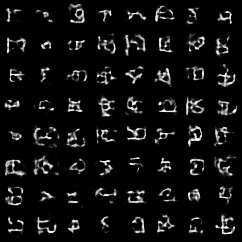}}
    \subcaptionbox{VampPriorVAE)}{\includegraphics[width=0.175\textwidth]{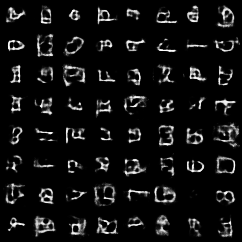}}
    \rulesep \rulesep
    \subcaptionbox{MixtureCSRAE}{\includegraphics[width=0.175\textwidth]{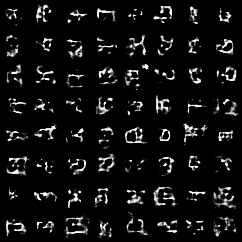}}
    \hfill \\
    \subcaptionbox{Caltech101}{\includegraphics[width=0.175\textwidth]{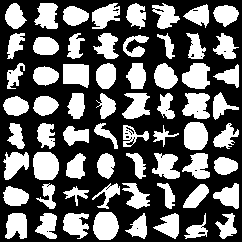}}
    \subcaptionbox{IWAE}{\includegraphics[width=0.175\textwidth]{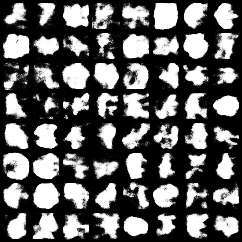}}
    \subcaptionbox{VampPriorVAE}{\includegraphics[width=0.175\textwidth]{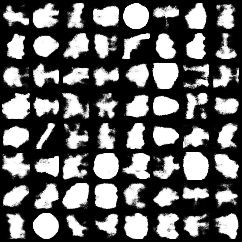}}
    \rulesep \rulesep
    \subcaptionbox{MixtureCSRAE}{\includegraphics[width=0.175\textwidth]{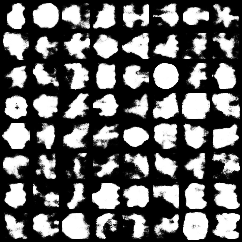}}
    \hfill \\
    \subcaptionbox{CIFAR10}{\includegraphics[width=0.175\textwidth]{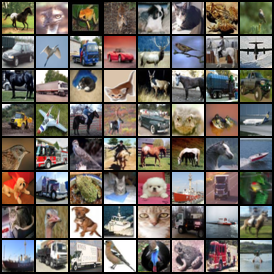}}
    \subcaptionbox{IWAE}{\includegraphics[width=0.175\textwidth]{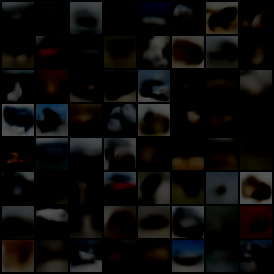}}
    \subcaptionbox{VampPriorVAE}{\includegraphics[width=0.175\textwidth]{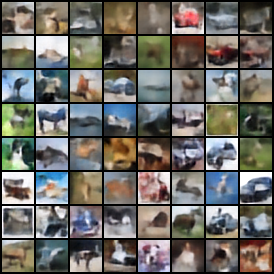}}
    \rulesep \rulesep
    \subcaptionbox{MixtureCSRAE}{\includegraphics[width=0.175\textwidth]{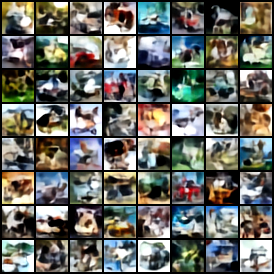}}
    \caption{Test samples (first column) and samples generated from IWAE (second column), VampPriorVAE (third column) and MixtureCSRAE (fourth column)}\label{fig:qual:samples}
    \hfill
\end{figure*}

\begin{figure*}[ht!]
    \centering
    \subcaptionbox{Dynamic MNIST}{\includegraphics[width=0.8\textwidth]{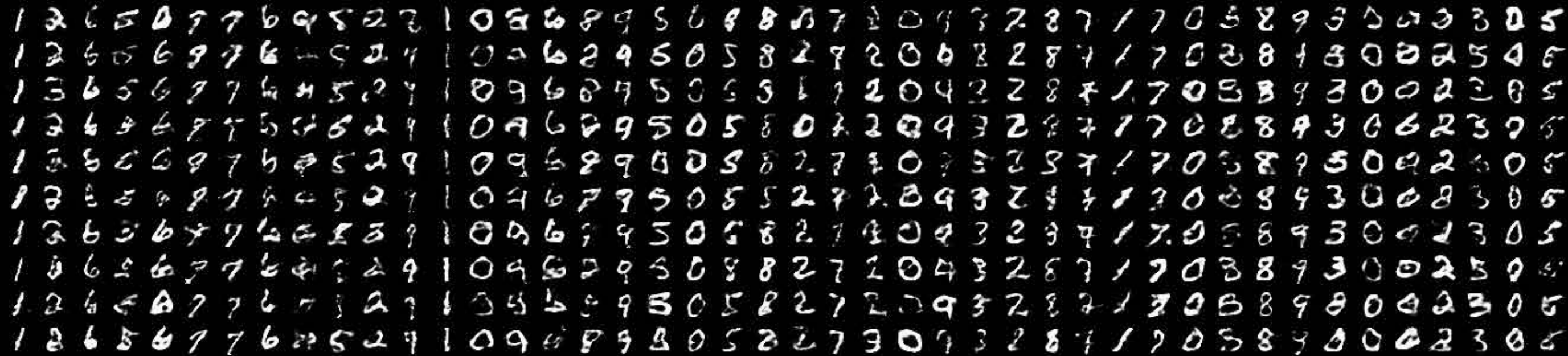}}
    \subcaptionbox{Caltech101}{\includegraphics[width=0.8\textwidth]{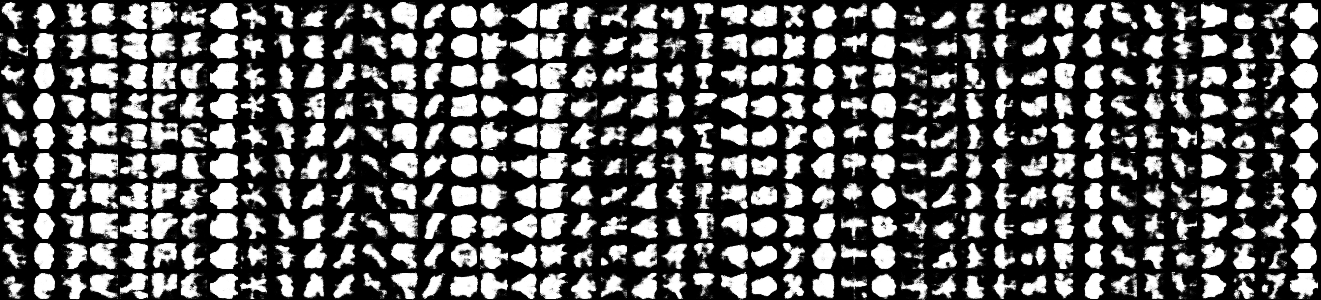}}
    \caption{Samples from MixtureCSRAE with a 100-component mixture of Gaussian prior trained with Dynamic MNIST (a) and Caltech101 (b). Each column of the samples represents samples coming from one component of the mixture of Gaussian.}\label{fig:qual:components}
\end{figure*}

\textbf{Qualitative results.}
We plotted both test samples next to its reconstruction in Figure~\ref{fig:qual:rec} for IWAE, VampPriorVAE and MixtureCSRAE. We notice with IWAE and VampPriorVAE that the reconstructions are often smooth even though the original samples had certain missing pixel and/or are distorted. With MixtureCSRAE the reconstructions seems to be more true to its original samples. Furthermore, IWAE seems to fail to reconstruct CIFAR samples. In comparison to VampPriorVAE, we noticed that MixtureCSRAE seems to be visually richer in contrast and sharper. We also visualized samples generated from the best performaning models in Figure~\ref{fig:qual:samples}. Similar to the reconstructions the samples of IWAE and VampPriorVAE are smoother and offer less diversity than MixtureCSRAE. Further, MixtureCSRAE samples for CIFAR10 appear to look sharper and richer in contrast. The diversity in samples is also shown in Figure~\ref{fig:qual:components} where we visualize samples from individual components of the Mixture of Gaussian prior. The samples are taken from best performing MixtureCSRAE with a Mixture of Gaussian prior of 100 components and trained with Dynamic MNIST and Caltech101. The samples show that each component covers a specific digit of MNIST or a specific shape of Caltech101, and also exemplifies the capability of each component covering diversity within class samples.
\subsection{Clustering} \label{subsec:clustering}
We compared the discriminative qualities of the model by using a k-Nearest Neighbors (kNN) on the latent samples of the test set of dynamic MNIST.
Table~\ref{tab:knn} show the results of CSRAE compared to the standard VAE (S-VAE) as well as related models, namely, the Gaussian mixture Variational Autoencoder
(GMVAE)~\cite{dilokthanakul2016deep}, the deep latent Gaussian mixture model
(DLGMM)~\cite{nalisnick2016approximate} and the Stick-Breaking Variational Autoencoder (SB-VAE)~\cite{nalisnick2017stick}.
For all models we used best top-1 accuracy on the validation for model selection and report the test accuracy for kNN with $k \in \{3, 5, 7\}$ in Table~\ref{tab:knn}.
We observed that settings that worked well for density estimation might not work well for kNN. Especially, we experienced an increase in classification error when using larger latent dimensions of $>20$. This decrease in performance can be attributed to the curse of dimensionality. As kNN uses Euclidean distance as default distance metric it becomes meaningless as latent dimension increases. Therefore, for all methods we only report latent dimensions of $[10, 20]$. 
Table~\ref{tab:knn} shows that without exception MixtureCSRAE outperforms all models when using latent samples for kNN. Our model improved kNN classification error by at least 2.65\% (Caltech101, $k=5$) to at most 25.12\% (Dynamic MNIST, $k=3$).
\begin{table}[t]
\centering
  \begin{tabular}{lcccccc}
        \toprule
        & \multicolumn{6}{c}{\textbf{Dataset}}\\
        & \multicolumn{3}{c}{Dynamic MNIST} & \multicolumn{3}{c}{Caltech101} \\
        \cmidrule(lr){2-4}\cmidrule(lr){5-7}
        Method & $k=3$ & $k=5$ & $k=10$ & $k=3$ & $k=5$ & $k=10$ \\
        \toprule
    VAE ($z=10$) & 5.20 & 4.78 & 4.52 & 41.35 & 40.10 & 40.26\\
    \midrule
    VAE ($z=20$) & 7.04 & 6.01 & 5.32 & 39.62 & 38.99 & 39.20\\
    \midrule
    VampPriorVAE~\cite{tomczak2018vae} &
        \multirow{2}{*}{4.92} &
        \multirow{2}{*}{4.79} &
        \multirow{2}{*}{4.90} &
        \multirow{2}{*}{41.74} &
        \multirow{2}{*}{39.90} &
        \multirow{2}{*}{40.91} \\ 
     ($z=10,k=400$) &&& &&& \\
    \midrule
    VampPriorVAE~\cite{tomczak2018vae} &
        \multirow{2}{*}{3.98} &
        \multirow{2}{*}{3.87} &
        \multirow{2}{*}{3.77} &
        \multirow{2}{*}{38.26} &
        \multirow{2}{*}{37.64} &
        \multirow{2}{*}{38.52}\\ 
    ($z=20,k=400$) &&& &&& \\
    \toprule
    CSRAE ($z=10$) & 4.22 & 3.93 & 3.81 & 41.39 & 39.73 & 40.32 \\ \midrule
    CSRAE ($z=20$) & 6.03 & 5.27 & 4.82 & 39.75 & 39.06 & 39.54 \\ \midrule
    MixtureCSRAE & 
        \multirow{2}{*}{3.91} &
        \multirow{2}{*}{3.79} &
        \multirow{2}{*}{3.78} &
        \multirow{2}{*}{\bf 37.88} &
        \multirow{2}{*}{37.62} &
        \multirow{2}{*}{36.75}\\
    ($z=10,k=400$) &&& &&& \\ \midrule
    MixtureCSRAE &
        \multirow{2}{*}{\bf 2.98} &
        \multirow{2}{*}{\bf 2.95} &
        \multirow{2}{*}{\bf 2.94} &
        \multirow{2}{*}{39.66} &
        \multirow{2}{*}{\bf 37.54} &
        \multirow{2}{*}{\bf 34.94}\\
    ($z=20,k=400$) &&& &&&\\
    \bottomrule
  \end{tabular}
  \caption{Average classification test error ($n=5$), lower is better. kNN classification was applied with different number of neighbours ($k=[3, 5, 10]$) on latent samples for dynamic MNIST and Caltech101. 
  }
  \label{tab:knn}
\end{table}
\subsection{Semi-supervised learning}
\label{subsec:ssl}
As introduced in Section~\ref{sec:ssl} we evaluate our semi-supervised approach with MNIST for digit classification as well as DISFA~\cite{mavadati2013disfa} and FERA2015~\cite{ZhangYCCRHLG14}. The results for MNIST is shown in Table~\ref{tab:ssl}(a). For all models we used the same architecture and optimization and report results on different number of labels used during training. For every model setting we report the average of five runs with different random seeds. When directly comparing VAE and CSRAE as well as VampPriorVAE and MixtureCSRAE our models improved average classification by at least 0.7\% (VAE vs. CSRAE) and at most 12.89\% (VampPriorVAE vs MixtureCSRAE). 

DISFA~\cite{mavadati2013disfa} and FERA2015~\cite{ZhangYCCRHLG14} are face datasets which evaluates learning of facial action units. 
These datasets have labelled facial action units according to the Facial Action Coding System (FACS)~\cite{ekman1997face} which define a set of facial muscle movements.
FACS allows to encode any anatomically possible facial expression and have shown wide applications in face and emotion recognition as well as mental health analysis. 
Most of the existing approaches for AU recognition are supervised and require a large number of facial action unit labels. However, FACS-based labelling is time-consuming and require expert knowledge. 
Table~\ref{tab:ssl} (b) and (c) shows the results w.r.t. average F1 score for DISFA (b) and FERA2015 (c). For each model setting and each fold we ran the experiments five times and report the average of it. We only compare our models to those of VampPriorVAE because VampPriorVAE consistently showed better performance than both VAE and IWAE. Further, we also report supervised results from a convolutional model with the same architecture as the encoder of our model (denoted as CNN in the results) and a Resnet18~\cite{he2016deep}. As both tables show, our results with MixtureCSRAE outperform the ones of VampPriorVAE. For DISFA, we show that with only 25\% of the labels we can achieve significant performance in comparison to the supervised equivalent (27.1 vs. 34.14). However, this seems to be also dependent on the dataset itself. While we could get closer to the supervised performance in the case of DISFA, we only could reach an F1 score of 27.11 in comparison to the equivalent supervised F1 score of 66.01. Another interesting observation is that for FERA2015 the F1 score decreases with increasing proportion of labels (30.56 vs. 27.11). As the individual AU performance differs greatly the average overall F1 performance is easily steered by very low or very high F1 score of a specific AU. In this case, increasing number of labels increased the individual F1 score of three out of five AUs, however, decreased overall average F1 score.
\begin{table*}[t]
    \centering
    \begin{subtable}[h]{0.31\textwidth}
        \resizebox{\textwidth}{!}{
            \begin{tabular}{l c c c}
                \toprule
                \multirow{2}{*}{Model} & \multicolumn{3}{c}{Number of labels used for training}\\
                \cmidrule{2-4}
                & $n=250$ & $n=1000$ & $n=4000$ \\
                \toprule
                VAE~\cite{kingma2013auto}& 4.89 & 3.78 & 3.55 \\
                VampPriorVAE~\cite{tomczak2018vae} & 4.73 & 3.85 & 3.15 \\
                \midrule
                CSRAE & 4.69 & 3.81 & 3.27 \\
                MixtureCSRAE & \textbf{4.12} & \textbf{3.73}& \textbf{2.75} \\ 
                \bottomrule
            \end{tabular}
        }
        \caption{Average  classification  test  error  ($n= 5$) for Dynamic MNIST, lower is  better.}
    \end{subtable}
    \hfill
    \begin{subtable}[h]{0.31\textwidth}
        \resizebox{\textwidth}{!}{
            \begin{tabular}{clcc}
                \toprule
                & \multirow{2}{*}{Model} & \multicolumn{2}{c}{Percentage of labels} \\
                & & \multicolumn{2}{c}{used for training} \\
                \cmidrule{3-4}
                & & $p=0.10$ & $p=0.25$ \\
                \toprule
                \parbox[t]{2mm}{\multirow{4}{*}{\rotatebox[origin=c]{90}{semi-sup.}}} & \multirow{2}{*}{VampPriorVAE~\cite{tomczak2018vae}} & \multirow{2}{*}{22.13} & \multirow{2}{*}{25.35} \\
                & & & \\
                & \multirow{2}{*}{MixtureCSRAE} & \multirow{2}{*}{\bf 25.35} & \multirow{2}{*}{\bf 27.10} \\ 
                & & & \\
                \bottomrule
                \parbox[t]{2mm}{\multirow{2}{*}{\rotatebox[origin=c]{90}{sup.}}} & CNN & \multicolumn{2}{c}{34.14}\\
                & Resnet18 & \multicolumn{2}{c}{45.76}\\
                \bottomrule
            \end{tabular}
        }
        \caption{Average F1 score from subject-independent 3-fold cross validation for DISFA (higher is better).}
    \end{subtable}
    \hfill
    \begin{subtable}[h]{0.31\textwidth}
        \resizebox{\textwidth}{!}{
            \begin{tabular}{clcc}
                \toprule
                & \multirow{2}{*}{Model} & \multicolumn{2}{c}{Percentage of labels}\\
                & & \multicolumn{2}{c}{used for training} \\
                \cmidrule{3-4}
                & & $p=0.10$ & $p=0.25$ \\
                \toprule
                \parbox[t]{2mm}{\multirow{4}{*}{\rotatebox[origin=c]{90}{semi-sup.}}} & \multirow{2}{*}{VampPriorVAE~\cite{tomczak2018vae}} & \multirow{2}{*}{23.07} & \multirow{2}{*}{25.86} \\
                & & & \\
                & \multirow{2}{*}{MixtureCSRAE} & \multirow{2}{*}{\bf 30.56} & \multirow{2}{*}{\bf 27.11} \\ 
                & & & \\
                \bottomrule
                \parbox[t]{2mm}{\multirow{2}{*}{\rotatebox[origin=c]{90}{sup.}}} & CNN & \multicolumn{2}{c}{66.01}\\
                & Resnet18 & \multicolumn{2}{c}{67.54}\\
                \bottomrule
            \end{tabular}
        }
        \caption{Average F1 score from subject-independent 3-fold cross validation for FERA2015 (higher is better).}
    \end{subtable}
    \caption{Classification results for Dynamic MNIST, DISFA and FERA2015. For all results, we report an average of five runs. With DISFA and FERA2015 we additionally average over the three subject-independent folds.}
    \label{tab:ssl}
\end{table*}

\subsection{Limitations}
There are several directions and limitations for further investigation, e.g., the usage of Mixture of Gaussian approximate posterior, the additional hyperameter that needs tuning and the possibilities of hierarchical modelling. We will briefly discuss it in the following.

\textbf{Mixture of Gaussian approximate posterior}
We investigated also not only using GMM for the prior, but also for the approximate posterior. Using GMM for the approximate posterior has been already proposed Nalisnick et al.~\cite{nalisnick2016approximate}. In~\cite{nalisnick2016approximate} the authors use a mixture of Gaussian approximate posterior with Dirichlet mixture weights. We have observed that a mixture of Gaussians for the approximate posterior did not improve expressiveness of the approximate posterior distribution. Rather, the optimization failed to assign meaningful weights. Concretely, during inference the approximate weights always assign high values to exactly one component, thus making the mixture weights obsolete. We can imagine several challenges why it might not work. The first challenge is that GMM reparametrization is non-trivial and requires either expensive marginalization or rejection sampling. Further, Dirichlet reparametrization relies on approximations which may impede optimization. We leave mixture of Gaussian approximate posterior for future investigation and focus only on GMM priors.

\textbf{Additional hyperparameter tuning} We introduced an additional hyperparameter for our (Mixture)CSRAE approach which accounts for the degree of regularization of the approximate posterior. We treated this regularization factor as hyperparameter and therefore, it increased the number of experiments that needed to be run during hyperparameter optimization. This increase in number of models required for hyperparameter optimization could be reduced by learning the regularization within the optimization process.

\textbf{Hierarchical modelling} Several works~\cite{tomczak2018vae,maaloe2019biva,vahdat2020nvae} have shown that hierarchical modelling of prior and approximate posterior can improve performance of auto-encoding models. We leave it for future work to explore hierarchical extensions of our Cauchy-Schwarz regularized autoencoder.
\section{Related Work}
The main focus of our work is on representation learning and density modeling in autoencoder-based generative models.
There are various works which address the expressiveness of posterior approximations and priors.
Coarsely, these streams of research can be categorized to (i) diagnosing the VAE framework, (ii) modified objective functions and (iii) more expressive prior and posterior approximations.

Many works have focused on identifying challenges in the VAE framework by dissecting the objective~\cite{hoffman2016elbo,zhao2017towards,alemi2018fixing}, and extending it to solve optimization issues~\cite{rezende2018taming,dai2018diagnosing}. With CSRAE, we argue that a simpler probabilistic objective is competitive for generative modelling. 

As was initially pointed out in~\cite{hoffman2016elbo} maximizing the ELBO might be not suitable in learning a good data representation. Many efforts have focused on resolving this problem by revising the ELBO. 
As a result, several works have been proposed to optimize a different bound or objective.
Hoffman et al.~\cite{hoffman2016elbo} introduce the aggregated posterior which is the expectation of the encoder over the data distribution.
They propose to improve the density estimation performance through minimizing the KL between aggregated posterior and prior.
However, the KL divergence between the aggregated posterior and prior cannot be calculated in closed form.
In contrast, CSRAE introduces a novel objective with closed-form approximation.
Further, looking at the VAE objective as a regularized auto-encoding one, different kinds of regularizer have been proposed.
The most prominent ones are adversarial loss as in Adversarial Autoencoders (AAEs)~\cite{makhzani2015adversarial} and
Wasserstein Autoencoders (WAEs)~\cite{tolstikhin2017wasserstein}.
Both AAEs and WAEs attempt to match the aggregated posterior and the prior, either by adversarial training or by minimizing their Wasserstein distance.
However, due to the use of deterministic encoders, there can be ``holes'' in the latent space which are not covered by the aggregated posterior which would result in poor sample quality~\cite{rubenstein2018latent}.
Within the framework of CSRAE, we still have a probabilistic encoder and have not encountered the challenges of only a small fraction of the total volume of the latent space being covered.

A different stream of works have focused on improving the expressiveness of approximate posterior and prior.
Works tending to the expressiveness of the posterior approximation include flow-based models such as normalizing~\cite{rezende2015variational}, auto-regressive~\cite{kingma2016improved} and Sylvester normalizing flows~\cite{van2018sylvester}.
These works transform the variational distribution into more complex ones by applying a sequence of invertible mappings.
Nalisnick et al.~\cite{nalisnick2016approximate} used GMMs as the approximate posterior for VAEs and improved the capacity of the original VAE.
Hoffman et al.~\cite{hoffman2016elbo} show that the prior plays an important role in the density estimation.
The standard Gaussian prior is usually used due to its efficiency and simplicity, however, this leads to overregularization of the latent variable and thus, a collapse of it. As a result, the performance w.r.t. density estimation is poor without any changes to the framework.
Other approaches modify a prior distribution, making it more complex than a original proposal: a Gaussian mixture of posterior approximations~\cite{tomczak2018vae,kuznetsov2019prior}, autoregressive priors~\cite{chen2016variational} or training a deterministic encoder and obtaining prior with a kernel density estimation~\cite{ghosh2019variational}.

Our method is closely related to the works trying to incorporate GMMs~\cite{nalisnick2016approximate, tomczak2018vae} to enable a richer posterior approximation. However, our work deviates from existing ones as we do not follow an objective that is not based on the variational Bayes approach.
\section{Conclusion}
In this paper, we proposed a new constrained optimization objective based on the Cauchy-Schwarz divergence to improve VAEs.
We followed the line of research that comparing the prior to the approximate posterior can result in a too restrictive posterior distribution and instead propose to match a mixture of Gaussians as approximate to a given prior. 
Further, we formulated an extended objective based on the Cauchy-Schwarz divergence which allows us to compute the divergence between mixtures of Gaussians analytically.
We showed empirically that using our objective we can increase the performance of the proposed generative model and improve discriminative abilities for clustering and semi-supervised tasks.

\bibliography{csrae}
\bibliographystyle{ieeetr}

\clearpage

\appendices
\section{Cauchy-Schwarz Regularized Autoencoder}\label{app:csrae}

In this section we show properties of the Cauchy-Schwarz divergence in Section~\ref{subsec:prop_csd}, the closed-form formulation of the Cauchy-Schwarz divergence for mixture of Gaussians in Section~\ref{app:subsec:csrae-deriv} and the analytical solution for our proposed MixtureCSRAE in Section~\ref{app:subsec:mcsrae-deriv}.
\subsection{Properties of the Cauchy-Schwarz divergence}\label{subsec:prop_csd}
\begin{itemize}
\item Symmetry:
\begin{align}
\begin{split}
D_{CS}(q(z) \parallel p(z|x)) =& \; - \log (\smallint q(z) p(z|x) dz) \\
& \; + 0.5 \log (\smallint q(z)^2 dz) \\
& \; + 0.5 \log (\smallint p(z|x)^2 dz)
\end{split}\\
=& \; D_{CS}(p(z|x) \parallel q(z))
\end{align}
\item $0 \le D_{CS} < \infty$, where $D_{CS}(p(\mb{x}) \parallel q(\mb{x})) = 0$ iff $p(\mb{x}) = q(\mb{x})$.
\begin{align}
D_{CS}(p(\mb{x}) \parallel q(\mb{x})) =& \; D_{CS}(p(\mb{x}) \parallel p(\mb{x})) \\
\begin{split}
=& \; \log (\smallint p(\mb{x}) p(\mb{x}) dz)\\
& \; + 0.5 \log (\smallint p(\mb{x})^2 dz)\\
& \; + 0.5 \log (\smallint p(\mb{x})^2 dz)
\end{split}\\
&= 0
\end{align}
\item Similar to KL, the Cauchy-Schwarz does not satisfy the triangle inequality and therefore cannot be classified as a metric. 
\end{itemize}
\subsection{Closed-form Cauchy-Schwarz divergence for mixture of Gaussians}\label{app:subsec:csrae-deriv}
Given the Gaussian PDF
\begin{align}
    \mathcal{N}(x; \mu, \sigma^2) = \frac{1}{\sqrt{2\pi \sigma^2}} e^{-\frac{(x-\mu)^2}{2\sigma^2}}
\end{align}
the product of two Gaussian PDFs is given by
\begin{align}
    \mathcal{N}(x; \mu_1, \sigma_1^2) \mathcal{N}(x; \mu_2, \sigma_2^2) = \mathcal{N}\bigg(\mu_1; \mu_2, \sqrt{\sigma_1^2 + \sigma_2^2}\bigg) \mathcal{N}(x; \mu_{12}, \sigma_{12}^2),
\end{align}
where
\begin{align}
    \mu_{12} = \frac{\sigma_1^{-2}\mu_1 + \sigma_2^{-2}\mu_2}{\sigma_1^{-2} + \sigma_2^{-2}}
\end{align}
and
\begin{align}
    \sigma^2 = \frac{\sigma_1^2 \sigma_2^2}{\sigma_1^2 + \sigma_2^2}.
\end{align}
This trick can be used to derive the analytical form of the Cauchy-Schwarz divergence for mixture-of-Gaussians.  Let
\begin{align}
    q(x) = \sum_{n=1}^N w_{n} \mathcal{N}(x | \mu_{n}, \sigma_{n}^2)
\end{align}
and
\begin{align}
    p(x) = \sum_{m=1}^M v_{m} \mathcal{N}(x | \nu_{m}, \tau_{m}^2)
\end{align}
be two mixture-of-Gaussian  (MoG) distributions with different parameters and different numbers of mixture components. The Cauchy-Schwarz divergence for a pair of MoGs can be derived as follows:
\begin{align}
    \textrm{D}_{\textrm{CS}}(q(x), p(x)) = -\underbrace{\log \bigg(\int q(x) p(x) dz\bigg)}_{\circled{1}} + 0.5 \underbrace{\log \bigg(\int q(x)^2 dx\bigg)}_{\circled{2}} + 0.5 \underbrace{\log \bigg(\int p(x)^2 dx\bigg)}_{\circled{3}}
\end{align}
We can use the product of Gaussian densities for each term, starting with \circled{1}
\begin{align}
    \log \bigg(\int q(x) p(x) dx\bigg) =& \; \log \bigg(\int \sum_{n=1}^N \sum_{m=1}^M  w_{n}  v_{m} \mathcal{N}(x | \mu_{n}, \sigma_{n}^2) \mathcal{N}(x | \nu_{m}, \tau_{m}^2) dx\bigg)\\
    =& \; \log \bigg(\sum_{n=1}^N \sum_{m=1}^M  w_{n}  v_{m} \int \mathcal{N}(x | \mu_{n}, \sigma_{n}^2) \mathcal{N}(x | \nu_{m}, \tau_{m}^2) dx\bigg)\\
    =& \; \log \bigg(\sum_{n=1}^N \sum_{m=1}^M  w_{n}  v_{m} \int \mathcal{N}\Big(\mu_n | \mu_m, \sqrt{\sigma_n^2 + \tau_n^2}\Big) \mathcal{N}(x | \mu_{n,m}, \sigma_{nm}^2) dx\bigg)\\
    =& \; \log \bigg(\sum_{n=1}^N \sum_{m=1}^M  w_{n} v_{m} \mathcal{N}\Big(\mu_n | \mu_m, \sqrt{\sigma_n^2 + \tau_n^2}\Big) \underbrace{\int \mathcal{N}(x | \mu_{n,m}, \sigma_{n,m}^2) dx}_{=1}\bigg)\\
    =& \; \log \bigg(\sum_{n=1}^N \sum_{m=1}^M  w_{n} v_{m} \mathcal{N}\Big(\mu_n | \mu_m, \sqrt{\sigma_n^2 + \tau_n^2}\Big)\bigg).
\end{align}
Similarly we can formulate \circled{2} as
\begin{align}
    \log \bigg(\int q^2(x) dx\bigg) =& \; \log \bigg(\int \sum_{n=1}^N \sum_{n'=1}^N  w_{n}  w_{n'} \mathcal{N}(x | \mu_{n}, \sigma_{n}^2) \mathcal{N}(x | \mu_{n'}, \sigma_{n'}^2) dx\bigg)\\
    =& \; \log \bigg(\sum_{n=1}^N \sum_{n'=1}^N  w_{n} v_{n'} \mathcal{N}\Big(\mu_n | \mu_{n'}, \sqrt{\sigma_n^2 + \sigma_{n'}^2}\Big)\bigg).
\end{align}
and \circled{3} as
\begin{align}
    \log \bigg(\int p^2(x) dx\bigg) =& \; \log \bigg(\int \sum_{n=1}^N \sum_{n'=1}^N  w_{n}  w_{n'} \mathcal{N}(x | \nu_{m}, \tau_{m}^2) \mathcal{N}(x | \nu_{m'}, \tau_{m'}^2) dx\bigg)\\
    =& \; \log \bigg(\sum_{n=1}^N \sum_{n'=1}^N  w_{m} w_{m'} \mathcal{N}\Big(\mu_m | \mu_{m'}, \sqrt{\tau_m^2 + \tau_{n'}^2}\Big)\bigg).
\end{align}
Putting it all together, we get
\begin{align}
    \begin{split}
         \textrm{D}_{\textrm{CS}}(q(x), p(x)) = & \; - \log \bigg(\sum_{n=1}^N \sum_{m=1}^M  w_{n} v_{m} \mathcal{N}\Big(\mu_n | \mu_m, \sqrt{\sigma_n^2 + \tau_n^2}\Big)\bigg) \\
         & \; + 0.5 \log \bigg(\sum_{n=1}^N \sum_{n'=1}^N  w_{n} v_{n'} \mathcal{N}\Big(\mu_n | \mu_{n'}, \sqrt{\sigma_n^2 + \sigma_{n'}^2}\Big)\bigg) \\ & \;+ 0.5  \log \bigg(\sum_{n=1}^N \sum_{n'=1}^N  w_{m} w_{m'} \mathcal{N}\Big(\mu_m | \mu_{m'}, \sqrt{\tau_m^2 + \tau_{n'}^2}\Big)\bigg)
    \end{split}
\end{align}
\subsection{Mixture Cauchy-Schwarz regularized autoencoder}\label{app:subsec:mcsrae-deriv}
Derivation of the objective function:
\begin{align}
\mathcal{L}_{\textrm{MixtureCSRAE}} = \underbrace{\E_{q_{\phi}(z|x)}[\log p_{\theta}(x|z)]}_{\circled{1}} - \underbrace{\lambda \textrm{D}_{\textrm{CS}}(q(z|x)\parallel p(z))}_{\circled{2}}\label{eq:LMixtureCSRAE}
\end{align}
\circled{2} :
\begin{align}
\begin{split}
    \lambda \textrm{D}_{\textrm{CS}}(q(z|x)\parallel p(z)) = & \; \lambda \bigg[- \log \Big(
    \frac{1}{K} \sum_{k}^{K} \mathcal{N}\big(\mu_{\phi} | \mu_{k, \psi}, \textrm{diag}(\sigma^2_{\phi} + \sigma^2_{k, \psi})\big)  + \log \mathcal{N}(\mu_{\phi} | \mu_{\phi}, \textrm{diag}(2\sigma^2_{\phi}))\\
    & \; \qquad + \log \Big(
    \frac{1}{K^2} \sum^K_{k, k'} \mathcal{N}(\mu_{k, \psi}|\mu_{k', \psi}, \textrm{diag}(2\sigma^2_{k',\psi}))
\Big)
\bigg]
\end{split}\\
\begin{split}
= & \; \lambda \bigg[- \log \Big(
    \sum_{k}^{K} \mathcal{N}\big(\mu_{\phi} | \mu_{k, \psi}, \textrm{diag}(\sigma^2_{\phi} + \sigma^2_{k, \psi})\big)\Big) + \log K - D \log (2\sigma_{\phi} \sqrt{\pi})\\
    & \; \qquad + \log \Big(
    \sum^K_{k, k'} \mathcal{N}(\mu_{k, \psi}|\mu_{k', \psi}, \textrm{diag}(2\sigma^2_{k',\psi}))
\Big) - 2 \log K
\bigg]
\end{split}\\
\begin{split}
= & \; - \lambda \log \Big(
    \sum_{k}^{K} \mathcal{N}\big(\mu_{\phi} | \mu_{k, \psi}, \textrm{diag}(\sigma^2_{\phi} + \sigma^2_{k, \psi})\big)\Big) +
    \lambda \log \Big(
    \sum^K_{k, k'} \mathcal{N}(\mu_{k, \psi}|\mu_{k', \psi}, \textrm{diag}(2\sigma^2_{k',\psi}))
\Big)\\
& \; + \lambda\log K - D \lambda \log (2\sigma_{\phi} \sqrt{\pi})
\end{split}
\end{align}
Putting \circled{2} back in \eqref{eq:LMixtureCSRAE}:
\begin{align}
    \begin{split}
        \mathcal{L}_{\textrm{MixtureCSRAE}} = & \; \E_{q_{\phi}(z|x)}[\log p_{\theta}(x|z)] + \lambda \log \Big(\sum_{k}^{K} \mathcal{N}\big(\mu_{\phi} | \mu_{k, \psi}, \textrm{diag}(\sigma^2_{\phi} + \sigma^2_{k, \psi})\big)\Big) \\
        & \; -
    \lambda \log \Big(
    \sum^K_{k, k'} \mathcal{N}(\mu_{k, \psi}|\mu_{k', \psi}, \textrm{diag}(2\sigma^2_{k',\psi}))
\Big) - \lambda\log K + D \lambda \log (2\sigma_{\phi} \sqrt{\pi})
    \end{split}
\end{align}
\section{Evaluation}
In order to reproduce all experiments, we describe the experimental setup as well as additional training procedures for facial action recognition for DISFA and FERA2015.

\subsection{Experimental setup}\label{app:subsec:exp_setup}
\begin{table}[t!]
    \centering
    \begin{subtable}[h]{0.4\textwidth}
        \centering
        \begin{tabular}{l l}
            \toprule
            \textbf{Parameters} & \textbf{Values} \\
            \toprule
            Batch size & 100 \\
            Latent dimension & [40, 128]\\
            Optimizer & Adam \\
            Adam: beta1 & 0.9\\
            Adam: beta2 & 0.999\\
            Adam: epsilon & 1$e$-8\\
            Adam: learning rate & 5$e$-4\\
            Training epochs & 400\\
            Warmup epochs & 100\\
            \bottomrule \\
        \end{tabular}
        \vspace{0.25cm}
        \caption{Hyperparameters common to each of the considered methods.}
        \label{tab:hyparameters_fixed}
    \end{subtable}
    \hfill
    \begin{subtable}[h]{0.59\textwidth}
        \centering
        \begin{tabular}{l c l}
            \textbf{Model} & \textbf{Parameters} & \textbf{Values} \\
            \toprule
            VAE & $\beta$ & $[0.1, 0.25, 0.5, 0.75, 1.0, 1.5, 2.0, 5., 10]$\\
            \midrule
            \multirow{2}{*}{IWAE} & $\beta$ & $[0.1, 0.25, 0.5, 0.75, 1.0, 1.5, 2.0, 5., 10]$\\
            & $n_{\textrm{iw}}$ & $[5, 50]$ \\
            \midrule
            \multirow{2}{*}{VampPriorVAE} & $\beta$ & $[0.1, 0.25, 0.5, 0.75, 1.0, 1.5, 2.0, 5., 10]$\\
            & $K$ & $[10, 100, 400]$\\
            \midrule
            CSRAE & $\lambda$ & $[0.1, 0.25, 0.5, 0.75, 1.0, 1.5, 2.0, 5., 10]$\\
            \midrule
            \multirow{2}{*}{MixtureCSRAE} & $\lambda$ & $[0.1, 0.25, 0.5, 0.75, 1.0, 1.5, 2.0, 5., 10]$\\
            & $K$ & $[10, 100, 400]$\\
            \bottomrule \\
        \end{tabular}
        \caption{Model hyperparameters. We allow five sweeps over a single hyperparameter for each model.}
        \label{tab:hyperparameters_var}
    \end{subtable}
    \caption{\textbf{Fixed and variable hyperparameters for unsupervised and semi-supervised learning}}
\end{table}
\begin{table}[t]
     \centering
    \begin{tabular}{l c c c c c c c}
        \toprule
        \multirow{2}{*}{\textbf{Dataset}} & \multirow{2}{*}{\textbf{Image dimension}} & \multirow{2}{*}{\textbf{Binarization}} & \multirow{2}{*}{\textbf{Decoder likelihood}} & \multicolumn{3}{c}{\textbf{Number of samples}} & \multirow{2}{*}{\textbf{Normalization range}}\\
        \cmidrule{5-7}
        & & & & Train & Validation & Test\\
        \toprule
        Static MNIST & $28 \times 28 \times 1$ & Static & Bernoulli & 45{,}000 & 5{,}000  & 10{,}000 & [0, 1]\\
        \midrule
        Dynamic MNIST & $28 \times 28 \times 1$ & Dynamic & Bernoulli & 45{,}000 & 5{,}000  & 10{,}000 & [0, 1]\\
        \midrule
        Omniglot & $28 \times 28 \times 1$ & Dynamic & Bernoulli & 23{,}128 & 1{,}217 & 8{,}070 & [0, 1]\\
        \midrule
        Caltech101 & $28 \times 28 \times 1$ & Dynamic & Bernoulli & 4{,}100 & 2{,}264 & 2{,}307 & [0, 1]\\
        \midrule
        CIFAR10 & $32 \times 32 \times 3$ & - & Discretized logistic & 45{,}000 & 5{,}000  & 10{,}000 & [-0.5, 0.5]\\
        \bottomrule\\
        \multirow{3}{*}{DISFA} & \multirow{3}{*}{$224 \times 224 \times 3$} & \multirow{3}{*}{-} & \multirow{3}{*}{Discretized logistic} & 78{,}488 & 8{,}721  & 43{,}605& \multirow{3}{*}{[-0.5, 0.5]}\\
        & & & & 78{,}489 & 8{,}721 & 43{,}604 & \\
        & & & & 78{,}488 & 8{,}721 & 43{,}605 & \\
        \bottomrule\\
        \multirow{3}{*}{BP4D+} & \multirow{3}{*}{$224 \times 224 \times 3$} & \multirow{3}{*}{-} & \multirow{3}{*}{Discretized logistic} & 112{,}479 & 12{,}498 & 72{,}811 & \multirow{3}{*}{[-0.5, 0.5]}\\
        & & & & 119{,}278 & 13{,}254 & 65{,}256 & \\
        & & & & 124{,}260 & 13{,}807 & 59{,}721 & \\
        \bottomrule\\
    \end{tabular}
    \caption{\textbf{Setups of all datasets used for evaluation.} Binarization is only used for Static MNIST, Dynamic MNIST, Omniglot, Caltech101, CIFAR10. For DISFA and BP4D+ we have three different sample sizes for train, validation and test due to 3-fold cross validation.}
    \label{tab:dataset_setup}
\end{table}
\label{table:arch}
\begin{table}[t!]
	\centering
	\begin{tabular}{c l l l}
	\toprule
	& \textbf{Dataset} & \multicolumn{2}{l}{\textbf{Architecture}}\\
	\toprule
	\parbox[t]{2mm}{\multirow{18}{*}{\rotatebox[origin=c]{90}{Unsupervised}}} & Pinwheel & $q_\phi(z|x)$ & FC 5, Softplus activation, FC 10, Softplus activation, FC 2 $\times$ latent dim.\\
    & & $p_\theta(x|z)$ & FC 10, Softplus activation, FC 5, FC 2 $\times$ 2 \\
	\cmidrule{2-4}
	& Static MNIST, & $q_\phi(z|x)$ & FC 300, ReLU act., FC 300, ReLU act., FC 2 $\times$ latent dim. \\
	& Dynamic MNIST, & $p_\theta(x|z)$ & FC 300, ReLU act., FC 300, ReLU act., FC 784, Sigmoid activation \\
	& Omniglot, & $p(z)$ & FC 784, ReLU act., FC 256, ReLU act., FC 2 $\times$ latent dim.\\
	& Caltech101 & &  \\
	\cmidrule{2-4}
	& CIFAR10 & $q_\phi(z|x)$ & Conv $128 \times 4 \times 4$ (Stride 2), ReLU act., Conv $256 \times 4 \times 4$ (Stride 2), ReLU act.,\\
	& & & Conv $254 \times 3 \times 3$ (Stride 1), \\
	& & & ResidualBlock[ReLU act., Conv $256 \times 3 \times 3$ (Stride 1), ReLU act., Conv $256 \times 1 \times 1$ (Stride 1)], \\
	& & & ResidualBlock[ReLU act., Conv $256 \times 3 \times 3$ (Stride 1), ReLU act., Conv $256 \times 1 \times 1$ (Stride 1)],\\
	& & & ReLU act., Conv $256 \times 1 \times 1$ (Stride 1), ReLU act., FC 2 $\times$ latent dim.\\
	& & $p_\theta(x|z)$ & FC 8192, Reshape (128, 8, 8), ReLU act., TransposeConv $256 \times 3 \times 3$ (Stride 1)\\
	& & & ResidualBlock[ReLU act., Conv $256 \times 3 \times 3$ (Stride 1), ReLU act., Conv $256 \times 1 \times 1$ (Stride 1)], \\
	& & & ResidualBlock[ReLU act., Conv $256 \times 3 \times 3$ (Stride 1), ReLU act., Conv $256 \times 1 \times 1$ (Stride 1)],\\
	& & & ReLU act., TransposeConv $128 \times 4 \times 4$ (Stride 2), ReLU act., TransposeConv $3 \times 4 \times 4$ (Stride 2)\\
	& & $p(z)$ & GatedDense 256, ReLU act., GatedDense 256, ReLU act., FC 2 $\times$ latent dim.\\
	\toprule
	\parbox[t]{2mm}{\multirow{18}{*}{\rotatebox[origin=c]{90}{Semi-supervised}}} & Dynamic MNIST & $h_x$ & FC 300, ReLU act., FC 300, ReLU act.\\
	& & $h_y$ & FC 256, ReLU act., FC 256, ReLU act.\\
	& & $q_\phi(y|x)$ & FC 300, ReLU act., FC 10\\
	& & $q_{\phi}(z | h_x, h_y)$ & FC 300, ReLU act., FC 2 $\times$ latent dim. \\
	& & $p_{\theta}(x|z, h_y)$ & FC 300, ReLU act., FC 300, ReLU act., FC 784, Sigmoid activation\\
	\cmidrule{2-4}
	& DISFA, & $h_x$ & Conv $32 \times 7 \times 7$ (Stride 4), BatchNorm 32, ReLU act., Conv $32 \times 7 \times 7$ (Stride 4), \\
	& BP4D+ & & BatchNorm 32, ReLU act., Conv $64 \times 4 \times 4$ (Stride 2), BatchNorm 64, ReLU act.,\\
	& & & Conv $64 \times 4 \times 4$ (Stride 2), BatchNorm 64, ReLU act., Conv $512 \times 4 \times 4$ (Stride 1), \\
	& & & BatchNorm 512, ReLU act.\\
	& & $h_y$ & FC 256, ReLU act., FC 256, ReLU act.\\
	& & $q_{\phi}(z | h_x, h_y)$ & FC 300, ReLU act., FC 2 $\times$ latent dim. \\
	& & $p_{\theta}(x|z, h_y)$ & TransposeConv $512 \times 1 \times 1$ (Stride 1), BatchNorm 512, ReLU act., \\
	& & & TransposeConv $64 \times 4 \times 4$ (Stride 1),  BatchNorm 64, ReLU act., \\
	& & & TransposeConv $64 \times 4 \times 4$ (Stride 2),  BatchNorm 64, ReLU act., \\
	& & & TransposeConv $32 \times 4 \times 4$ (Stride 2),  BatchNorm 32, ReLU act., \\
	& & & TransposeConv $32 \times 6 \times 6$ (Stride 4),  BatchNorm 32, ReLU act., \\
	& & & TransposeConv $16 \times 6 \times 6$ (Stride 4),  BatchNorm 16, ReLU act., \\
	& & & TransposeConv $3 \times 1 \times 1$ (Stride 1)\\
	\bottomrule\\
	\end{tabular}
	\caption{\textbf{Model architectures.} The architectures are common to the evaluation for unsupervised (density estimation, kNN clustering) and semi-supervised experiments.}
	\label{tab:model_architecture}
\end{table}
In our evaluation, we fix all hyperparameters except one latent dimensions which are all listed in Table~\ref{tab:hyparameters_fixed}. Model specific hyperparameters can be found in Table~\ref{tab:hyperparameters_var}.

Static MNIST has fixed binarization of image pixels~\cite{larochelle2011neural} where dynamic MNIST with dynamic binarization during training~\cite{salakhutdinov2008quantitative}.
Omniglot~\cite{lake2015human} contains $1,623$ hand-written characters from $50$ various alphabets with $20$ images per class.
Caltech 101 Silhouettes (Caltech101) is a dataset with silhouette images of $101$ object classes.
The silhouettes are black polygons of the corresponding objects on a white background.
Similar to dynamic MNIST, dynamic binarization was applied for both Omniglot and Caltech101 during train and test.
CIFAR10~\cite{krizhevsky2009learning} consists of 60,000 colour images of $32 \times 32$ in 10 classes, with 6000 images per class. All the image data sets images were either normalized with pixels between $0$ and $1$ or $-0.5$ and $0.5$. Empirically, we found that normalizing $-0.5$ and $0.5$ for a discretized logistic likelihood performed better than leaving it normalized between $0$ and $1$. These settings including binarization, likelihood and number of samples for train, validation and test can be found in Table~\ref{tab:dataset_setup}. FERA 2015 contains about 140{,}000 images from 41 subjects. Intensities are annotated for 5 AUs. DISFA contains per frame intensity annotations from 27 subjects of 8 AUs on a 6-point ordinal scale. For DISFA and BP4D we normalized the images using facial landmarks. 
We extract the locations of the eyes from facial images in each dataset using facial landmark annotations. We used the two facial points to calculate the average points in each dataset to define a reference frame. For that, a similarity transform was employed as in ~
\cite{ZengCTCX15,ZhaoCTCZ15}. The final image size is $256 \times 256$. For training we randomly cropped to images of size $224 \times 224$, and applied horizontal mirroring and random rotation for data augmentation. For validation and testing, we only center cropped images to size $224 \times 224$.

For each type of evaluation (density estimation, kNN clustering, semi-supervised learning) we used the same architecture for our proposed approach and all comparison methods. 
This is to ensure fair comparisons between all models. 
For grayscale image datasets with image sizes of $28 \times 28$, we used MLP based architectures whereas for RGB image datasets we used residual (CIFAR10) and convolutional (Fera, DISFA) architectures.
The architectures used are depicted in Table~\ref{tab:model_architecture}.
\subsection{Semi-supervised learning for facial action unit recognition}\label{app:subsec:ssl_face}
\begin{figure*}[t!]
\centering
\subcaptionbox{Fera2015}{\includegraphics[width=0.35\linewidth]{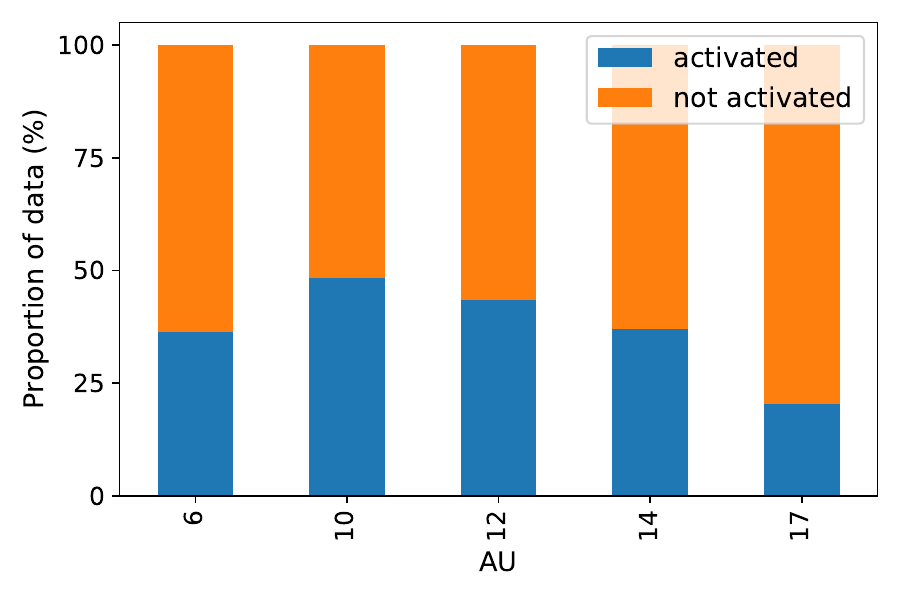}\label{fig:fera_dist_lab}}%
\subcaptionbox{DISFA}{\includegraphics[width=0.35\linewidth]{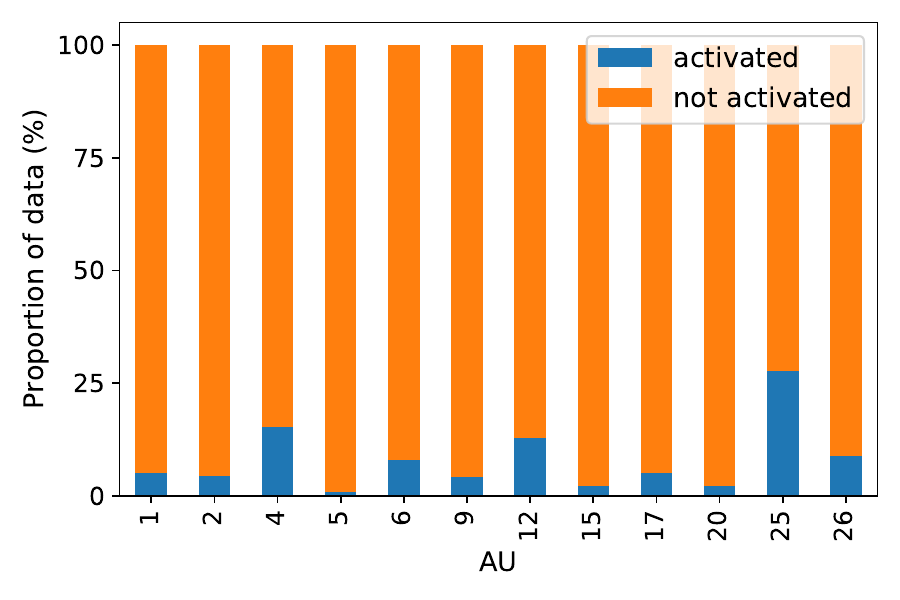}\label{fig:disfa_dist_lab}}
\caption{Facial action unit (AU) label distribution for FERA2015 and DISFA with respect to the amount of data instances with certain AU being activated or not.}
\label{fig:lab_dist}
\end{figure*}
For facial action unit recognition, we applied iterative label-balanced batches to the training to deal with imbalanced datasets. Further, similar to \cite{KingmaMRW14} we also used a two phase training scheme.

\textbf{Two-phase semi-supervised optimization} We first trained DISFA and FERA2015 in an unsupervised fashion without any labels involved. For unsupervised training, we used the same architectures as for the semi-supervised training. After unsupervised training, we used the pretrained weights of the encoder and decoder for training with labels. During evaluation we observed an improvement in overall performance when the encoder and decoder were already pretrained.

\textbf{Iterative label-balanced batches} One of the difficulties in recognizing facial action units is due to the imbalance in label distribution. AU activation occurs rarely and varies considerably among subjects as visualized in Figure~\ref{fig:lab_dist}. To tackle this challenge, we introduce iterative balanced batches to deal with the data imbalance during training. We generate batches that are balanced with respect to each action unit by undersampling from the majority class. Since the proportion of positive (activated) samples differ with each action unit, we generate balanced batches with respect to each action unit in an iterative manner. A drawback is that we are removing information that may be valuable. This could lead to underfitting and poor generalization to the test set.

Tables~\ref{tab:ssl_disfa} and~\ref{tab:ssl_fera} show detailed results w.r.t. to F1 score for DISFA and FERA2015. In particular, it shows average F1 scores for individual AUs as well as the overall average F1 score for each model.

\begin{table}[t]
    \centering
    \begin{tabular}{c cc cccc}
        \toprule
        \multirow{2}{*}{AU} & \multicolumn{2}{c}{Supervised} & \multicolumn{4}{c}{Semi-supervised}\\
         & & & \multicolumn{2}{c}{$p=0.1$} & \multicolumn{2}{c}{$p=0.25$}\\
        \cmidrule(l){4-5} \cmidrule(l){6-7}
        & Resnet18 & CNN & VampPriorVAE & MixtureCSRAE & VampPriorVAE & MixtureCSRAE\\
        \toprule
        1 & 2.44 & 2.13 & 0.79 & \bf 3.75 & 2.52 & 3.67\\
        \midrule
        2 & 26.82 & 2.93 & 0.88 & 4.37 & 3.59 & \bf 4.75\\
        \midrule
        4 & 56.66 & 43.39 & 13.64 & 14.69 & 27.90 & \bf 29.26 \\
        \midrule
        6 & 35.67 & 23.60 & 27.51 & \bf 32.20 & 26.26 & 30.47 \\
        \midrule
        9 & 27.86 & 21.35 & 7.28 & 12.74 & 15.52 & \bf 16.39\\
        \midrule
        12 & 66.82 & 56.23 & 54.84 & 58.81 & 56.01 & \bf 60.57\\
        \midrule
        25 & 88.28 & 87.63 & 54.71 & \bf 57.32 & 54.49 & 54.22\\
        \midrule
        26 & 39.62 & 35.83 & 17.39 & \bf 17.81 & 16.50 & 17.46\\
        \bottomrule
        Avg. & 45.76 & 34.14 & 22.13 & 25.21 & 25.35 & \bf 27.10\\
        \bottomrule
    \end{tabular}
    \caption{Average F1 score from subject-independent 3-fold cross validation for DISFA (higher is better).}
    \label{tab:ssl_disfa}
\end{table}
\begin{table}[t]
    \centering
    \begin{tabular}{c cc cccc}
        \toprule
        \multirow{2}{*}{AU} & \multicolumn{2}{c}{Supervised} & \multicolumn{4}{c}{Semi-supervised}\\
         & & & \multicolumn{2}{c}{$p=0.1$} & \multicolumn{2}{c}{$p=0.25$}\\
        \cmidrule(l){4-5} \cmidrule(l){6-7}
        & Resnet18 & CNN & VampPriorVAE & MixtureCSRAE & VampPriorVAE & MixtureCSRAE\\
        \toprule
        6 & 77.05 & 72.50 & 17.17 & 24.68 & \bf 25.81 & 19.30\\
        \midrule
        10 & 77.86 & 78.88 & 34.77 & \bf 60.46 & 33.04 & 39.10\\
        \midrule
        12 & 85.98 & 80.59 & 32.50  & 36.32 & 29.39 & \bf 39.65\\
        \midrule
        14 & 44.50 & 53.18 & 21.60 & 28.16 & 26.90 & \bf 32.08\\
        \midrule
        17 & 50.30 & 9.25 & 3.43 & 2.41 & 7.11 & \bf 9.25\\
        \bottomrule
        Avg. & 67.54 & 66.01 & 23.24 & 30.56 & 23.07 & \bf 27.87\\
        \bottomrule
    \end{tabular}
    \caption{Average F1 score from subject-independent 3-fold cross validation for FERA2015 (higher is better).}
    \label{tab:ssl_fera}
\end{table}

%



\ifCLASSOPTIONcaptionsoff
  \newpage
\fi

\end{document}